\documentclass[onecolumn,9pt,nofootinbib]{article}
\pdfoutput=1
\usepackage[T1]{fontenc}
\usepackage[utf8]{inputenc}
\usepackage{authblk}
\usepackage{amsmath}
\usepackage{graphicx}
\usepackage{color}
\usepackage{url}

\def\s{\sigma}
\def\dd{\delta}

\newcommand{\Ssf}{\mathsf{S}}
\newcommand{\Hcal}{\mathcal{H}}

\newcommand{\Lcal}{\mathcal{L}}

\begin{document}
\title{Maximum entropy models capture melodic styless}
\author[*]{Jason Sakellariou}
\author[**]{Francesca Tria}
\author[**]{Vittorio Loreto}
\author[*]{Fran\c{c}ois Pachet}
\affil[*]{SONY CSL, Paris, France \\ \& Sorbonne Universit\'es, UPMC Univ Paris 06, UMR 7606, LIP6, F-75005, Paris, France}
\affil[**]{Department of Physics, Sapienza University of Rome, Rome \\ \& ISI Foundation, Turin, Italy}

  \maketitle

\begin{abstract}
We introduce a Maximum Entropy model able to capture the statistics of melodies in music. The model can be used to generate new melodies that emulate the style of the musical corpus which was used to train it. Instead of using the $n-$body interactions of $(n-1)-$order Markov models, traditionally used in automatic music generation,  we use a $k-$nearest neighbour model with pairwise interactions only. In that way, we keep the number of parameters low and avoid over-fitting problems typical of Markov models.  We show that long-range musical phrases don't need to be explicitly enforced using high-order Markov interactions, but can instead emerge from multiple, competing, pairwise interactions. We validate our Maximum Entropy model by contrasting how much the generated sequences capture the style of the original corpus without plagiarizing it. To this end we use a data-compression approach to discriminate the levels of borrowing and innovation featured by the artificial sequences. The results show that our modelling scheme outperforms both fixed-order and variable-order Markov models. This shows that, despite being based only on pairwise interactions, this Maximum Entropy scheme opens the possibility to generate musically sensible alterations of the original phrases, providing a way to generate innovation.
\end{abstract}

\section*{Introduction}

Many complex systems exhibit a highly non-trivial structure that is difficult to capture with simple models. Several biological systems form networks of interacting components (neurons, proteins, genes, whole organisms) whose collective behavior is characterized by a complex mosaic of correlations among the different components. Arguably, the ultimate biological origin of purely intellectual constructs such as language or music, should allow us to look at them from a similar point of view,  i.e., as complex networks of interacting components.  In both cases, one would suspect  that essential features of their complexity arise from high-order combinatorial  interactions. However, a number of works in recent years have shown that  models based on \emph{pairwise} interactions alone capture most of the correlation structure of some biological systems \cite{schneidman2006weak,lezon2006using,bialek2007rediscovering,Weigt2009,mora2010maximum, bialek2012statistical} and even English words \cite{StephensBialek2007}. In this paper we extend this idea to the field of music.

One of the most popular strategies for algorithmic music composition is that of Markov chains (see for example \cite{curtis1996computer}). In this setting music is seen as a sequence of symbols (these can be notes, chords, etc.) and is generated probabilistically by assigning conditional probabilities on those symbols given the preceding ones. In order to imitate the style of an existing musical corpus one can \emph{learn} these probabilities by counting the number of occurrences of substrings of symbols, or $k$-grams, in that particular corpus.   In order to capture the long-range structure of musical phrases, high-order Markov models must be used, i.e., probabilities are conditioned on $(k-1)$-grams for some large $k$.  Such an approach can lead to serious over-fitting issues: the number of actually represented $k$-grams in a musical corpus is usually orders of magnitude smaller than their total potential number, which is exponential in $k$.  Typical musical corpora contain a few hundred notes when the total number of different pitches is a few tens. When this is the case, probabilities for patterns longer than bi-grams (i.e., pairs of symbols)  is estimated with very poor accuracy.  For example, J.S. Bach's first violin Partita contains $1910$ notes when  the size of the alphabet, i.e., the number of distinct notes used, is $33$. In that case the number of bi-grams in the corpus and the  total number of possible bi-grams are comparable, and so the estimation of bi-gram probabilities should be fairly accurate. This is, however, not true for $k-$grams with $k$ greater than $2$. Although this is just a particular example, pieces with  a number of notes greater than quadratic to the alphabet size would be unnaturally long and are seldom found in music.

On the other hand, music is governed by a very rich and non-trivial set of rules, which may seem highly arbitrary and combinatorial. For instance in western tonal music, certain triplets of notes (such as C, E and G) are considered valid chords whereas the vast majority of three-notes combinations are rarely, if at all, used. Moreover, hardly any of these rules seems to have a fundamental character as they vary considerably across different cultures and epochs. At first sight it may seem impossible to capture the rich structure of a musical piece by a model that only takes into account pairwise information.  Work in biological systems, however, has suggested that this need not be true \cite{bialek2007rediscovering}. In this paper, we show that, for musical data, enforcing pairwise consistency across different time-distances restricts the space of solutions enough for higher-order patterns to emerge. That way, we capture long range musical patterns while avoiding the over-fitting issues of high-order models. This approach cannot be implemented as an extension of Markov models,  and a  different framework is needed. This framework is provided by the  \emph{Maximum Entropy principle} \cite{jaynes1957information}. Maximum entropy models consistent with pairwise correlations are variations of the  Ising or Potts models of statistical mechanics (see for instance \cite{baxter2007exactly}), which have a long and rich history as theoretical models for statistical order and phase transitions.   These models belong to the large family of \emph{Probabilistic Graphical Models}, which offer a very general framework for modeling statistical dependencies. We show here that our model can be used for generating sequences that mimic some aspects of the musical style of  a given corpus.

\section*{Results}
\label{seq:results}

\subsection*{The Model}
\label{sec:model}

Music has many dimensions (melody, harmony, rhythm, form, sound, etc) which renders realistic models extremely complicated. In this paper we focus on monophonic pitch sequences, for simplicity. A pitch sequence is a sequence of integer variables $\{s_1, \dots, s_N\}$ encoding note pitches ordered as they appear in a real melody but disregards other information about duration, onset, velocity etc. The variables take values from some finite alphabet $s_i \in \{1,\dots,q\}$ which are indices of types of musical pitches. In our setting we are given an initial pitch sequence, called the \emph{corpus} throughout the  paper, of which we want to learn the style.

\begin{figure}[!htb]
  \begin{center}

    \centerline{\scalebox{0.38}{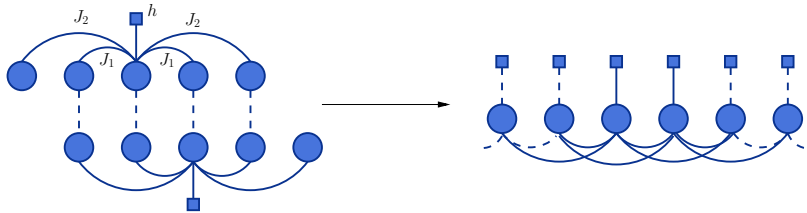}}
  \end{center}
  \caption{{\bf The Graph Representation.} Section of a  graph representing the factorization of the distribution     (\ref{eq:maxEntProb}) for $K_{\text{max}}=2$. The topology of the graph reflects the way variables interact in the Hamiltonian. Interaction potentials (edges in the graph) and local fields (square nodes) are connected to variables (circle nodes)  according to~(\ref{eq:maxEntProb}). A model is built by taking the union of smaller modules shifted by one variable, avoiding duplicate edges. Each module models the way each note depends on its local context (refer to the method section.}
  \label{fig:graph}
\end{figure} 

We will use here the The \emph{Maximum Entropy} principle \cite{jaynes1957information}, looking for the distribution $P$ that maximizes the entropy $S = - \sum_{\{s_i\}} P(s_1, \dots, s_N) \log P(s_1, \dots, s_N)$, and reproduces the corpus frequencies of single notes and of pairs of notes at distance $k$, with $k=1,\dots, K_{\text{max}}$  (refer to the section methods for details).
Using Lagrange multipliers to solve the above constraint optimization problem we obtain the following Boltzmann-Gibbs distribution:

\begin{equation}
P(s_1, \dots, s_N) = \frac{1}{Z} \exp\left(\sum_{i=1}^N h(s_i) + \sum_{k=1}^{K_{\text{max}}} \sum_{\substack{i,j \\ j-i=k}} J_k(s_i,s_j) \right) 
\label{eq:maxEntProb}
\end{equation}
\noindent   where the partition function 
\begin{equation}
Z \equiv \sum_{s_1}\sum_{s_2}\cdots \sum_{s_N} \exp\left(\sum_{i=1}^N h(s_i) + \sum_{k=1}^{K_{\text{max}}} \sum_{\substack{i,j \\ j-i=k}} J_k(s_i,s_j) \right)
\label{eq:partFunc}
\end{equation}

guaranties that the distribution is normalized. We will refer to the $h$'s as the local fields and to the $J_k$'s as the interaction potentials.  Adopting a statistical physics point of view,  these quantities can be thought as external fields acting on the  variables on one hand and interactions between variables on the other hand. The Hamiltonian then gives the energy of the system by summing the contribution of all the above terms. According to distribution~(\ref{eq:maxEntProb}), sequences with low energy have larger probability. Therefore, the effect of the above potentials is to bias the  probabilities of different sequences of notes.  It is important to note that we are interested in the statistics of notes and pairs of notes independently of their exact position in the sequence. The single-note marginals should be all equal and the pair marginals should depend only on the distance between notes (refer again to the methods section).
Actually, in music, position matters as the choice of notes depends strongly on a particular
context. Here however we
chose to focus on a \emph{translation-invariant} model for simplicity, i.e., one where 
single and double 
point statistics would look the same on every neighbourhood of size of order $O(K_{\text{max}})$.
This leads to a model that is constructed by repeating a basic module which models note
relations locally (see Fig.~\ref{fig:graph}
and method section for more details).

 In the methods section we present a method for choosing the values of these potentials in such a way as to make note frequencies of the model consistent with the ones found in a musical corpus.

Once the potentials have been found one can generate new pitch sequences by sampling from distribution~(\ref{eq:maxEntProb}). This can be simply done  by the Metropolis Algorithm~\cite{metropolis1953equation}.  We start from a random sequence. Then we repeat the following procedure: we pick a note at random, compute its probability conditioned by its neighbours, 
given by eq.~(\ref{conditionalProb}) and draw a value from this probability.
In practice we found that a number of Monte Carlo steps equal to $T_{\text{MC}}=10 N$ is 
sufficient to achieve good results, according to criteria  described in the 
next sections.

Musical style imitation is a difficult concept to grasp and formalize. However, most musicians would agree that it involves two things: creatively rearranging existing material from the musical corpus one wants to imitate and developing the existing ideas into new ones that resemble the original material. We shall call these two activities \emph{imitation} and \emph{innovation}.
Concerning imitation, we don't look for an arbitrary reshuffling of substrings of notes, or \emph{melodic patterns} as we will call them.  In the new sequence, these patterns must follow ``naturally'' one another just as in the corpus.  Concerning innovation, the new material cannot be random. One could argue that it should be statistically consistent with the corpus, by emphasizing the same notes and note pairs for example. A model that aims at imitating a given musical style should therefore be able to  create music using existing melodic patterns and invented new ones in a way consistent with the corpus.  We claim that our model fulfils the above criteria.

% Results and Discussion can be combined.

\begin{figure}[!htb]
  \begin{center}
   \centerline{\includegraphics[width=0.6\columnwidth]{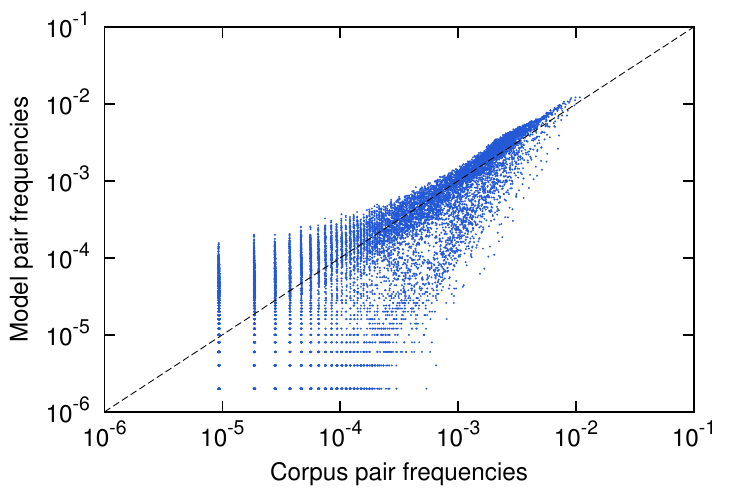}}
  \end{center}
    \caption{{\bf Model VS Corpus pair frequencies.} The Corpus ones are from the corpus \cite{WeimarJazzDatabase} (see Section S3 in the SI for additional information).
      The model frequencies come from a $N=200000$ sequence generated by a $K_{\text{max}}=10$ model trained on the above corpus.}
\label{fig:scatter}
 \end{figure}

\subsection*{Pairwise Correlations}
\label{subsec:pairCorrel}

We first look at what the model should do by construction: reproduce the correct single and double note frequencies. Fig.~\ref{fig:scatter} shows a scatter plot for  pair frequencies of the corpus versus the ones generated by our model. Single note frequencies yield very similar scatter plots, just with much fewer points hence we decided not to show them.  
For this particular example we used as a corpus  the content of the Weimar Jazz Database\cite{WeimarJazzDatabase} consisting of 257 transcriptions of  famous Jazz improvisations.
For more information about this corpus, as well as other corpora used in the experiments
throughout this paper, we refer the reader to Section S3 of the SI.
There is very good agreement for the more frequent pairs and, as expected,  small probabilities are reproduced less accurately. There is a fraction of note pairs that are  under-represented in generated sequences. It seems to be difficult for the basic Monte Carlo algorithm to access them. However, the great majority of note-pair probabilities are very well aligned with the corpus, as shown in Fig.~\ref{fig:scatter}.

\begin{figure}[!htb]
  \begin{center}
    \centerline{\includegraphics[width=0.7\columnwidth]{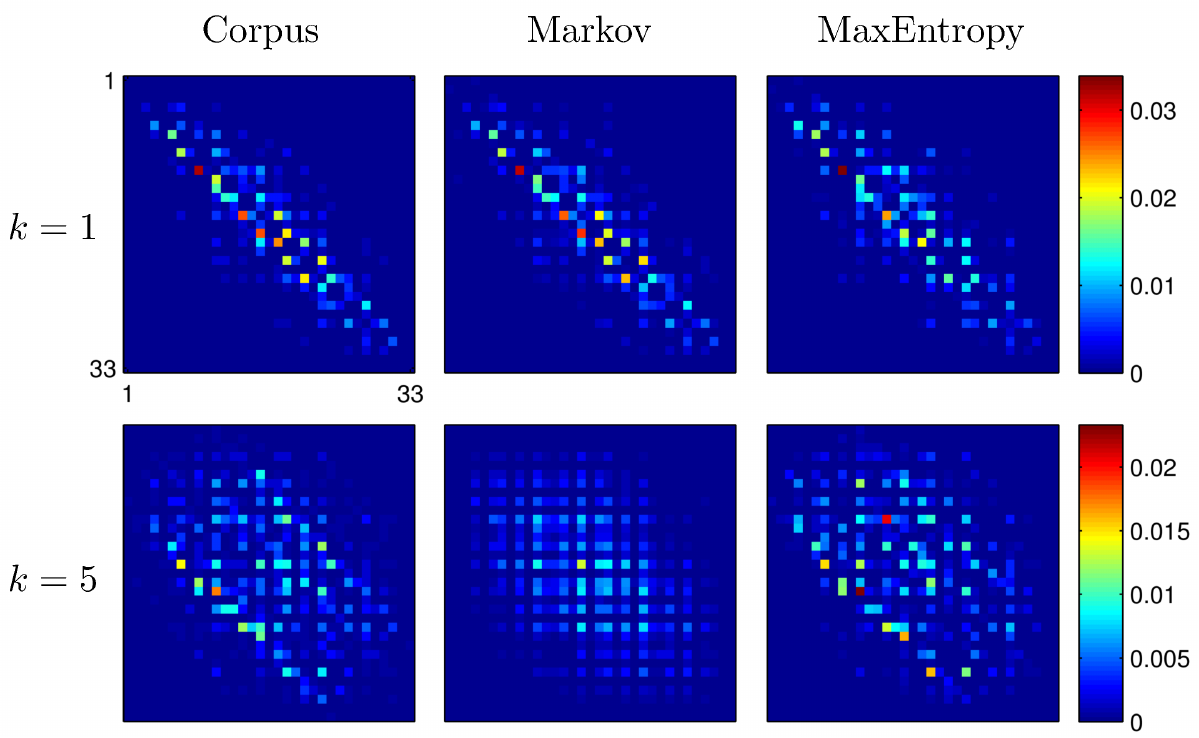}}
  \end{center}
    \caption{{\bf Matrices of pair frequencies.} Color-maps representing matrices obtained by counting pair frequencies (see eq.~(\ref{eq:freqs2})). From left to right: the original sequence by J.S. Bach (see Section S3 of the SI), a first-order Markov model and our Maximum Entropy model. Top row: pairs at distance $k=1$, bottom row : the same at $k=5$.}
    \label{fig:freqMatr}
\end{figure}

To better appreciate what the model does, it is informative to look at pair frequencies for different distances separately.  Fig.~\ref{fig:freqMatr} shows color-maps of matrices given by eq.~(\ref{eq:freqs2}) for $k=1$ and $k=5$ for three cases: the original sequence, here Partita No. 1 in B minor BWV 1002 by  Johann Sebastian Bach part II double (see Section S3 of the SI),  a first-order Markov model and our Maximum Entropy model.   The Markov model, by construction, reproduces perfectly the frequencies of  neighboring notes ($k=1$). However it fails to do so at greater distances. In  the bottom row we see that the particular information contained in the $k=5$ matrix of the corpus is almost completely lost for the Markov case. The Maximum Entropy model, however,  performs equally well in both cases. Here we used a model with $K_{\text{max}}=10$ so the training will make sure to select a set of potentials that better reproduce the pair frequencies for all distances up to 10.  The reason for using a first-order Markov model for comparison with our model is that they have comparable sample complexities. First-order Markov models have $O(q^2)$ parameters, where $q$ is the alphabet size, and so they need roughly the same amount of samples, or greater, to be trained accurately. Our model  has $O(K_{\text{max}}q^2)$ which makes the two models comparable. In contrast, a $K_{\text{max}}$\nobreakdashes-order Markov model has a sample complexity of $O(q^{(K_{\text{max}}+1)})$. We don't have to show the corresponding matrices for a high-order Markov model as it would reproduce correctly  the pair frequencies at all distances by trivially copying the whole corpus.  To summarize, our model maintains a quadratic sample complexity by using only pairwise constraints but is able to achieve long range consistency by combining multiple such constraints.

\begin{figure}[!htb]
  \centering
  \includegraphics[width=0.6\linewidth]{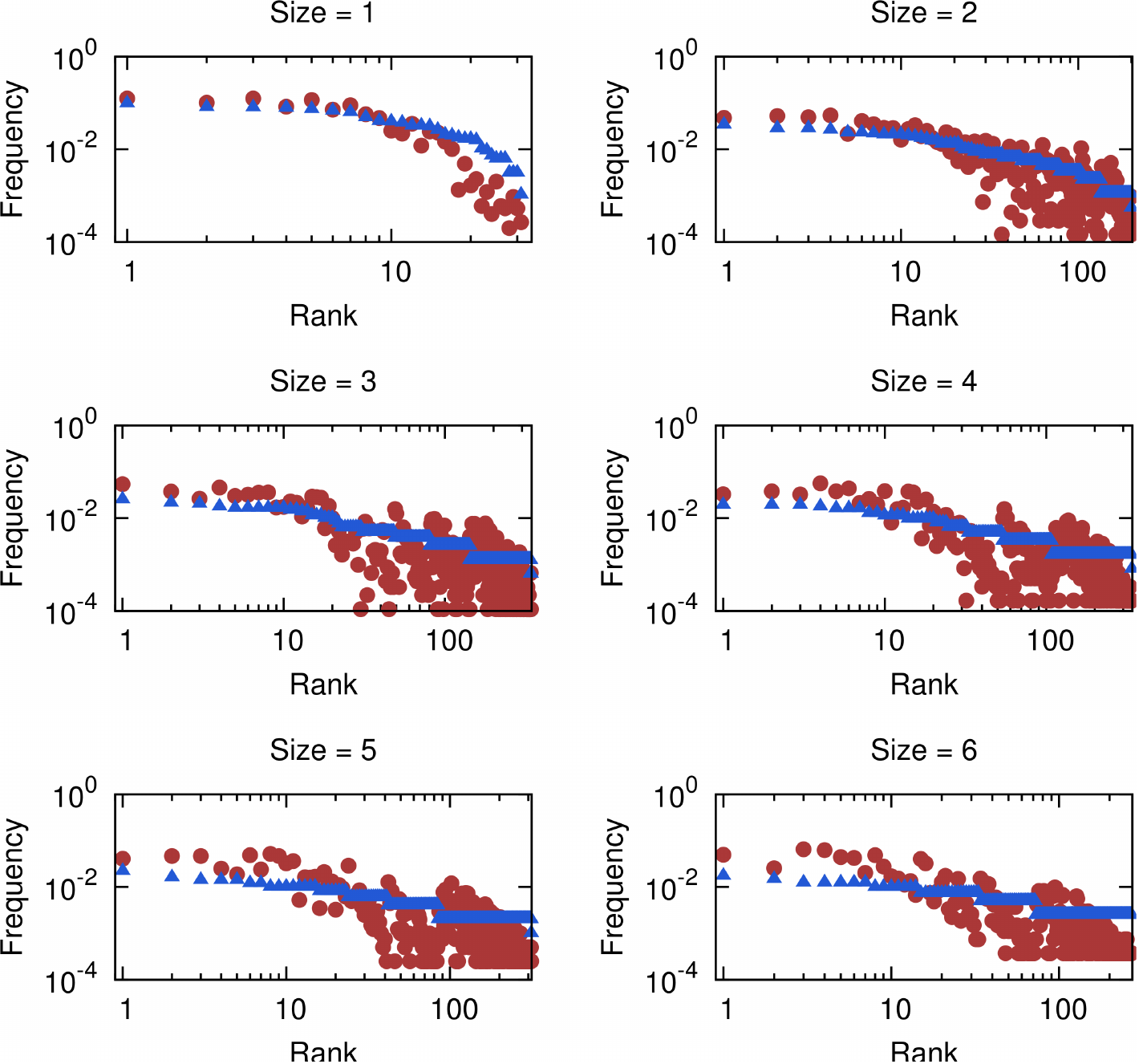}
  \caption{{\bf Frequency-rank plots for pattern frequencies.} In blue, all patterns appearing in the corpus of sizes 1 to 6 are ranked according to their frequency. Here the corpus is the Bach Partita used previously (see Section S3 of the SI). Then, the same patterns are located in a $N=15000$ sequence generated from a model trained accordingly. In red we plot their frequencies in the generated sequence but using the same order as before in order to compare with the corresponding frequencies in the corpus.}
\label{fig:rankFreq}
 \end{figure}

\subsection*{Higher-order Patterns}
\label{subsec:highOrdPatt}

Pairwise correlations are explicitly enforced into the model by using pairwise interaction  potentials, so it is not very surprising to correctly reproduce them. However, real music contains recognizable patterns, i.e., subsequences, of size greater than two. A music generating model should capture these higher-order patterns as well. Our model succeeds in reproducing higher-order patterns by combining multiple pairwise constraints. Fig.~\ref{fig:rankFreq} shows a series of frequency-rank plots for patterns of different sizes. Precisely, a $N=15000$ long sequence is generated from a $K_{\text{max}}=30$ model trained on a J.S. Bach partita (see Section S3 of the SI). Then an exhaustive search returns all patterns of sizes one up to six which are also present in the corpus. Finally we compute their frequencies,  in the corpus and in the generated sequence, and plot them in decreasing order with respect to the corpus probabilities. In order to have comparable results we normalize the  frequencies within the set of common patterns since our model also creates new  patterns which are not present in the corpus (see next Section and Section S4 of the SI for details on this feature ). The plots show that our model is indeed able to capture high-order patterns and to reproduce them with fairly consistent probabilities. 

\subsection*{Borrowing and Innovation}
\label{subsec:innovation}

\noindent  We have seen above that though the model only explicitly enforces pairwise constraints, longer melodic patterns can also be generated (see Fig.~\ref{fig:rankFreq}). We now make a step further and evaluate how well the generated melodic sequences capture the style of the original corpus without plagiarizing it. Two extremes are competing: borrowing and innovation. On the one hand the patterns generated can be identically appearing in the original corpus. The length of these patterns determines how much the generated sequence is {\em borrowing} from (or plagiarizing) the original corpus. If these patterns were too long one would trivially recognize the style, lacking in this way of originality. On the other hand {\em innovation} would imply that not all melodic patterns in the generated sequences are identical to ones found in the corpus. For example, if in a particular corpus the following patterns are  present $abx$, $axc$ and $xbc$ with $x$ substituting any character except $c$,$b$ and $a$  respectively, then the pattern $abc$ is likely to emerge although it was never part of the corpus. We call this feature innovation as it resembles the basis of all  creative processes: combining features of existing ideas to form new ones. In order to quantify the interplay between borrowing and innovation we consider  suitable observables through which we evaluate the goodness of the artificial sequences generated with three methods: our Maximum Entropy model, the fixed-order Markov model and the variable-order Markov model.

We have already discussed fixed-order Markov models. In these models $k$-grams are continued according to conditional probabilities estimated from the corpus. In this case it is very difficult for instance to control the Longest Common Substring between the artificial sequence and the corpus. In Fig.~S1 of the Supporting Information it can be seen  that the Longest Common Substring (LCS) for fixed-order Markov grows very fast with $K_{\text{max}}$, here the order of the model, leading to total plagiarism. \emph{Variable-order Markov models} (VO) were invented \cite{begleiter2004prediction} to circumvent this problem. Like in  fixed-order Markov models each note is drawn from a distribution conditioned in the preceding $k-$gram, but this time the size $k$ can vary at each step according to some  criterion. A simple implementation that resolves the plagiarism problem is to use at  each step of the generation the largest $k<K_{\text{max}}$ that leads to more than, say, 3 different continuations, where $K_{\text{max}}$ here is a maximal order chosen by the user.  That makes plagiarism exponentially unlikely. This version of the variable-order Markov model has been successfully used in  \cite{pachet2003continuator}. In these models the LCS quickly saturates to a particular value. Beyond this, changing $K_{\text{max}}$ doesn't have any effect since a much smaller $k$ is always selected. As for our Maximum Entropy Model, Fig.~S1 of the SI shows that the  Longest Common Substring grows roughly linearly with $K_{\text{max}}$.

\begin{figure}[!htb]
    \centering
    \includegraphics[width=0.55\linewidth]{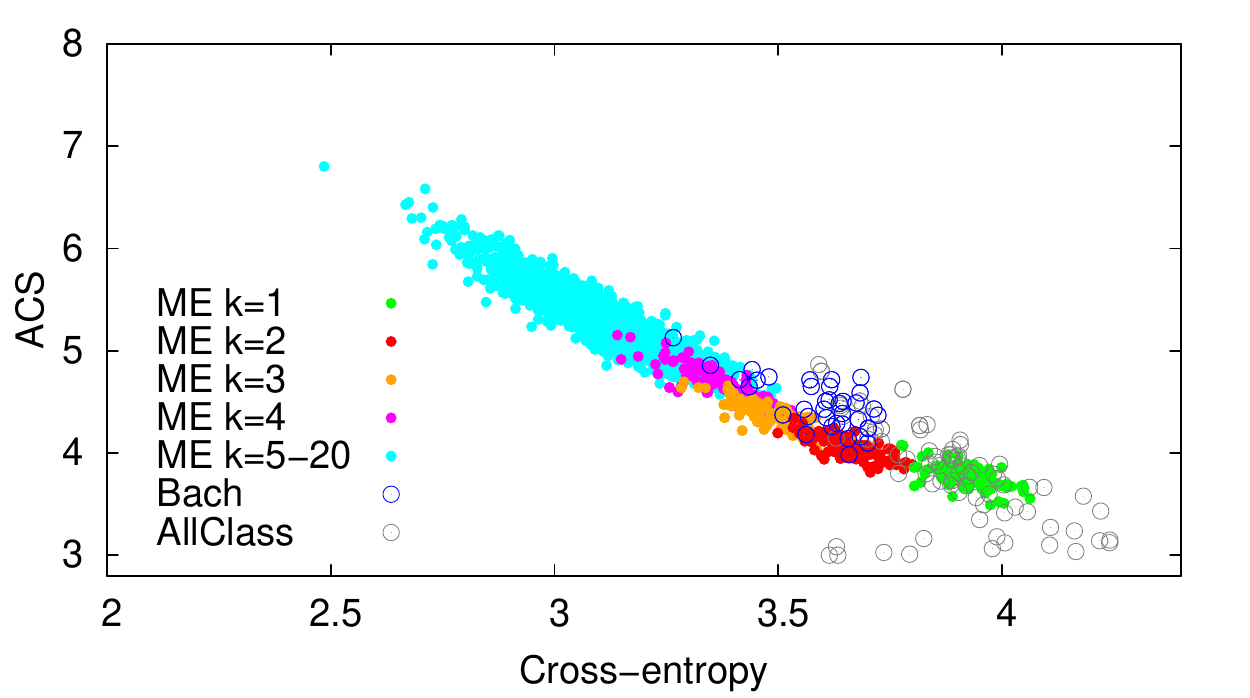}              
     \includegraphics[width=0.55\linewidth]{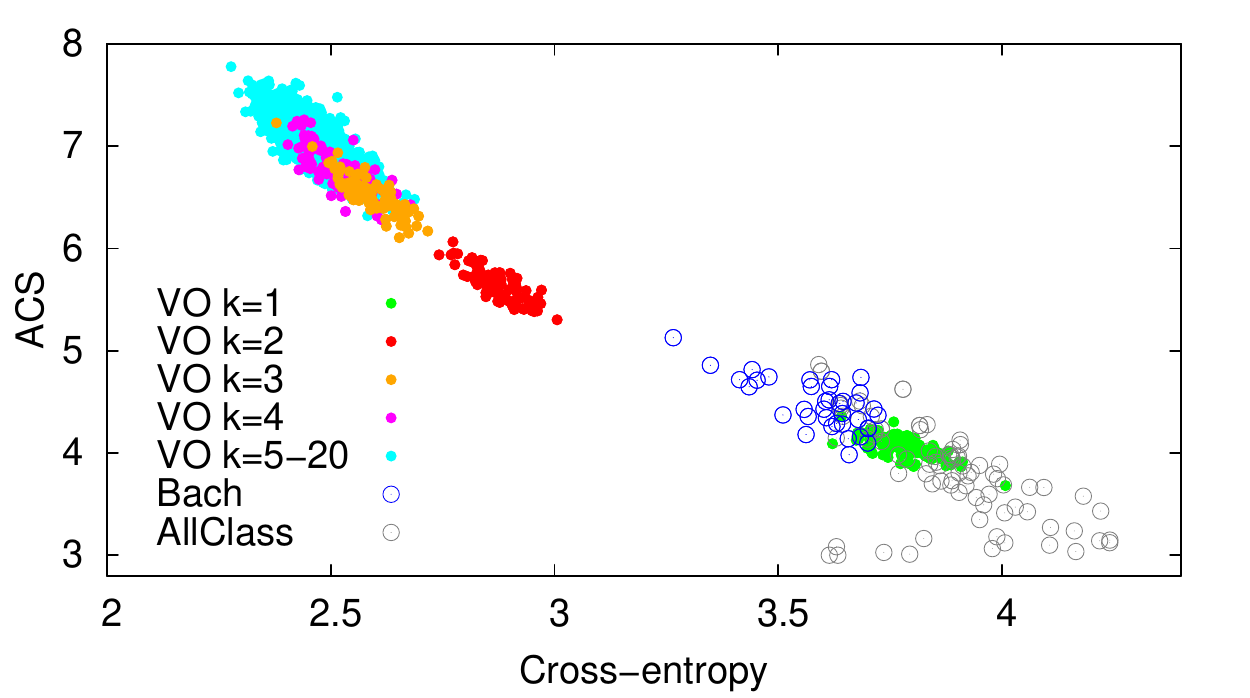}   
     \includegraphics[width=0.55\linewidth]{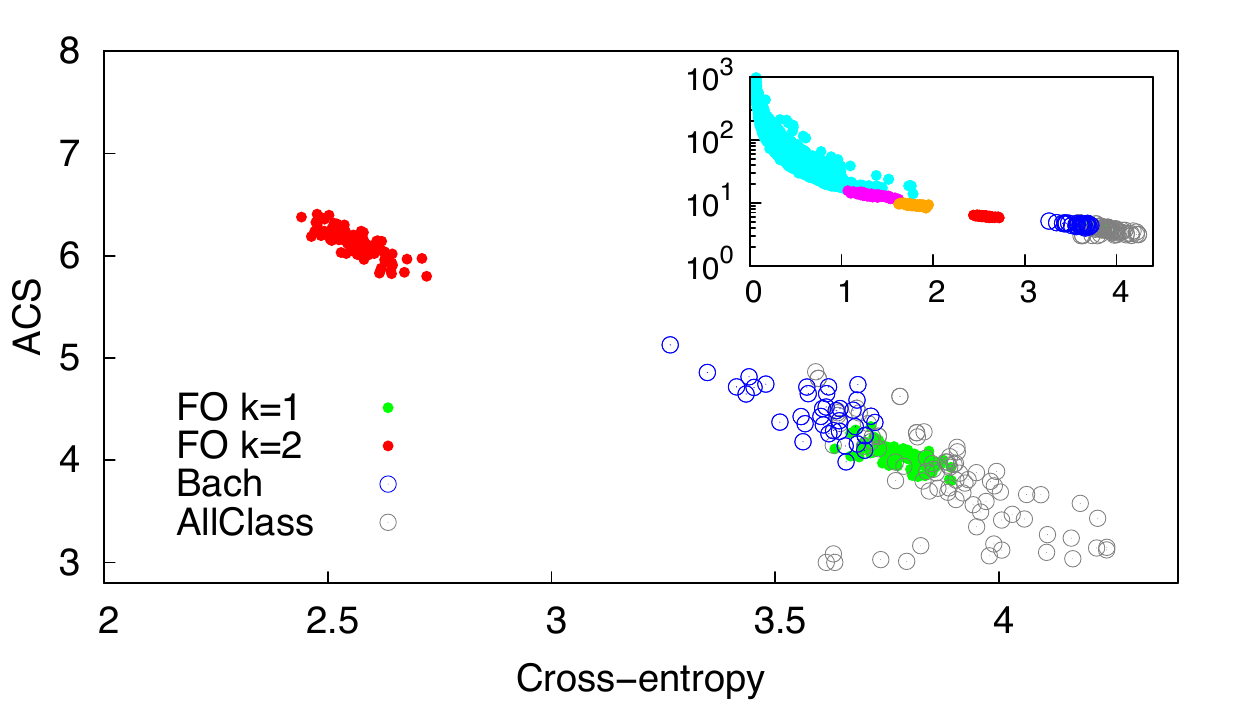}                         
    \caption{{\bf Borrowing vs. similarity.} This figure reports the Average Common Substring (ACS) vs. the values of the cross-entropies for all the artificial sequences generated with the Maximum Entropy (ME) model (top), the variable-order (VO) Markov model (middle) and the fixed-order (FO) Markov model (bottom). Everything is computed with respect to the sequence of J.S. Bach's first violin Partita (see Section S3 of the SI). Filled circles correspond to the artificial sequences. Colors code for the values of $K_{\text{max}}$ in each different model. In addition in each panel  the empty circles reports the same quantities for  other original sequences of Bach (represented with blue circles) and other classical authors - Beethoven, Schumann, Chopin, Liszt and Albeniz - (AllClass represented with grey circles). Note that the main panel for FO is truncated at values of ACS equal to 8, while the complete plot is shown in the inset.}
    \label{fig:Innovation_vs_similarity}
\end{figure}

In order to quantify the ability of the different models to capture the style of a corpus,
we look at the similarity between any artificially generated sequence and the corresponding training corpus. First, in order to avoid detecting similarities/dissimilarities due to two
sequences being in the same/different tonalities, we switched to a different point of view.
In equal-temperament music, one can create different versions of the same melody 
by transposing it to a different tone. We want to treat these transposed versions
as being equivalent. 
So, instead of encoding absolute pitch for each note, we code intervals between successive notes.
In other words, we only consider the difference between successive terms in every sequence of 
pitches.
The new sequence contains the same information as the old one, except for an additive constant
which contains tonality information.
Then we need a method capable of capturing some sort of similarity between these sequences of
intervals.

Given two sequences $\cal{A}$ and $\cal{B}$ we adopt the notion of cross-complexity as a measure of the remoteness between them, the cross-complexity being the algorithmic version of cross-entropy~\cite{chaitin_1977,cerra_2012}. We define the cross-complexity (though here we shall refer to it as cross-entropy) of a sequence  $\cal{B}$ with respect to another sequence $\cal{A}$ as the amount of bits needed to specify $\cal{B}$ once $\cal{A}$ is known. We follow a refined version of the data-compression approach introduced in~\cite{bcl_2002,zippers_2005}, that was shown to be successful in authorship attribution and corpora classification~\cite{zippers_2005}. We use in particular the LZ77 compressor~\cite{LZ77} and we scan the $\cal{B}$  sequence looking for existing matching sub-sequences only in $\cal{A}$ and we code each matching as in the usual LZ77 algorithm. 
In this way we estimate the cross-entropy of each artificially generated sequence with respect to the corresponding sequence of the original corpus. Details about the usage of data compression techniques to estimate the cross-entropy between two sequences are reported in the Supporting Information.

The cross-entropy described so far helps us in quantifying the similarity of the artificially generated sequences with the original corpus. The smaller the value of the cross-entropy the larger is the similarity. Now a small-value of the cross-entropy may be due to a  genuine {\em stylistic} similarity between the two sequences. In this case the artificial sequence looks like the original corpus without plagiarizing it, i.e., without borrowing large subsequences of the corpus itself. On the other hand, a small cross-entropy may be due to the presence of large chunks of the original corpus in the artificial sequence. To discriminate between these two cases we look at another observable, namely the {\em Average Common Substring} (ACS) between the artificial sequence and the corpus. The ACS is also computed using the data-compression technique described above. Given two sequences $\cal{A}$ and $\cal{B}$ and all the substrings found by the LZ77 algorithm while parsing $\cal{B}$ for matching in $\cal{A}$, the ACS is defined as the average length of all the matches found. We compute the ACS of each artificially generated sequence with respect to the corresponding sequence of the original corpus. Small ACS implies an important degree of innovation in the artificial sequence, while a large value of ACS implies a high degree of borrowing.

Overall, while the cross-entropy informs us about how statistically similar is the artificially generated sequence to the original corpus, the ACS tells us about the degree of borrowing from the original corpus. Fig.~\ref{fig:Innovation_vs_similarity} illustrates the results of this analysis performed using J.S. Bach's first violin Partita in B minor as the original corpus.  It is clear that the Maximum Entropy model is the only one able to capture values of similarity and level of borrowing comparable to those of other corpora from Bach (blue circles). Both variable-order and fixed order Markov models either feature a low level of borrowing but large dissimilarities (green filled circles for $k=1$) or high similarity with the original corpus but large values of ACS, i.e., a high level of borrowing. From this perspective the Maximum Entropy model features an optimal balance between similarity with the original corpus with a level of borrowing comparable to that found between different original pieces of Bach. Similar results for other original corpora (from Beethoven, Schumann, Chopin, Liszt) are reported in the Supporting Information. A similar analysis is also reported in the SI, where LCS is considered instead of ACS, bringing to similar results. Finally, in order to let the reader to evaluate "musically" the artificial pieces generated by our Maximum Entropy model, we provide, as audio Supporting Information,  a series of wav files including original pieces (both classical and jazz) and the corresponding artificially generated ones (see for details Section S7 of the textual Supporting Information).

\section*{Discussion}

\label{sec:conclusion}

We presented a Maximum Entropy model that captures pairwise correlations between notes in a musical sequence, at various distances.  The model is used to generate original sequences that mimic a given musical style. The particular topology  of this model (see Fig.~\ref{fig:graph}) leads to the emergence of high-order patterns, despite the pairwise nature of the information used, which in turn has the benefits of a quadratic, in the alphabet size, sample complexity.   Moreover, the absence of high-order constraints and their substitution by  multiple pairwise constraints, gives our model more freedom to create new melodic material that imitates the musical style of a given corpus. 

One key questions arising when proposing a specific algorithm to generate artificial musical sequences with the {\em style} of a given corpus is how to validate the results, i.e., to provide a quantitative account of how much the sequences generated by the model are similar to the original corpus without plagiarizing it. To this end we considered two specific observables to quantify the levels of borrowing and innovation in the generated sequences. Based on data-compression techniques, these two observables allow to claim that Maximum Entropy models, like the one proposed here, outperform both fixed-order and variable-order Markov models in providing musically sensible alternative realizations of the style of a given corpus. 

Finally graphical models like the one proposed here offer a general framework for modeling statistical dependencies. Work is in progress to extend this modeling scheme in order to account for other aspects of music, such as rhythm, polyphony and expressivity. The general idea is the same: additional information (e.g., note durations) can be captured  by additional variables coupled with \emph{pairwise} interactions. In that way, one can keep the quadratic sample complexity, while avoiding over-fitting, and create models able to make musically sensible generalizations of the ideas found in the corpus.

\section*{Methods}
\subsection*{Model details}

We are interested in reproducing the corpus frequencies of single notes and of pairs of notes at distance $k$, 
\begin{eqnarray}\label{eq:freqs1}
f(\s) &\equiv& \frac{1}{N} \sum_{i=1}^{N} \dd(\s,s_i)  \\ 
f_k(\s,\s') &\equiv& \frac{1}{N-k} \sum_{\substack{i,j \\ j-i=k}} \dd(\s,s_i)\dd(\s',s_j) 
\label{eq:freqs2}
\end{eqnarray}
with $k=1,\dots, K_{\text{max}}$. 
In the above formulas $\dd(\cdot,\cdot)$ is the Kronecker delta symbol. The sums run over the whole corpus.

It is crucial to note that the above quantities have no dependence on the position within the sequence, i.e., they do not depend on the indices $i$ and $j$. The first quantity in eq.~(\ref{eq:freqs1}) represents  the frequency of notes, regardless of the position at which they appear. The quantity in eq.~(\ref{eq:freqs2}) represents the frequency of co-occurrence of pairs of notes, again regardless of position, depending however on the distance between  the variables.

We look for the distribution (or probabilistic model) $P$ that maximizes the entropy $S = - \sum_{\{s_i\}} P(s_1, \dots, s_N) \log P(s_1, \dots, s_N)$ (Maximum Entropy principle) and that satisfies:
\begin{eqnarray}
f(\s)&=&P(\s)    \\ \label{eq:constraints}
\text{and} \; \; \; \; f_k(\s,\s')&=&P_k(\s,\s') \nonumber
\end{eqnarray}
where in the right hand side we have the marginals of the model's distribution
\begin{equation}
P(\s)  \equiv  P_i(s_i=\s) = 
\sum_{\{s_k | k \neq i\}} P(s_1, \dots, s_N), \;\; \;\;\; \forall i  
\label{marginal1}
\end{equation}

\begin{equation}
\begin{split}
P_k(\s,\s') \equiv  P_{ij}(s_i =\s,s_j = \s')  = \phantom{blablablablablabla} \\ 
\phantom{blablablablabla} \sum_{\{s_l|l \neq i,j\}} P(s_1, \dots, s_N) \;\;\;\;\;\; \forall i,j: j-i=k
\end{split}
\label{marginal2}
\end{equation}

\noindent Note that a model can have any desired length, i.e., any number of notes. However, the interactions between variables extend to a maximum length given by $K_{\text{max}}$ which is usually much smaller than the total length of the model. Moreover, interactions for same-distance variables repeat themselves along the graphical model, as described earlier. The interaction graph is therefore highly regular and can be seen as constructed from some basic module. This module is composed of one variable node, its local field and all its first-neighbours. It has size  $2 K_{\text{max}} +1$ and contains a copy of all the parameters of the model, i.e., one local field $h$ and two copies of each interaction potential $J_k$. Each module models the way each note depends on its local context. In order to build a bigger model we take the union of two such modules shifted  by one variable, avoiding duplicate edges, as shown in Fig.~\ref{fig:graph}. This procedure is then repeated a number of times until the desired  total number of variables $N$ is reached. That creates a translation invariant model, except for regions of size $K_{\text{max}}$ on the borders, which will have a negligible  effect since $K_{\text{max}} \ll N$.

\begin{figure}[!htb]
  \begin{center}
%%\centerline{\includegraphics[width=0.9\columnwidth]{figs/data.pdf_t}}  
  \centerline{\scalebox{0.50}{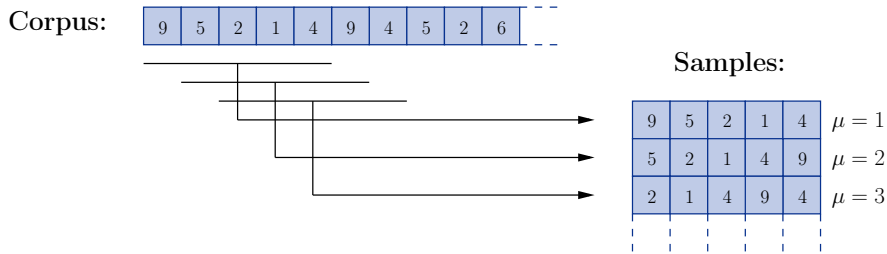}}
  \end{center}
  \caption{{\bf Training data.} The corpus, a sequence of type indices, is segmented by  overlapping substrings of size $2 K_{\text{max}} +1$ (here $K_{\text{max}}=2$). These samples provide the information needed to train the basic module of our model, which describes the way a variable depends on its local context, i.e., on $K_{\text{max}}$ variables to its left and to its right.}
\label{fig:data}
  \end{figure}

\noindent The above picture of a longer model built by the union of simpler modules allows us to simplify the task of choosing the values of the potential in eq.~(\ref{eq:maxEntProb}) by realizing that we need to infer the potentials only for one such module.

\subsection*{Parameter learning}
\label{sec:training}

We use a \emph{Maximum Likelihood Estimation} (MLE) of the model parameters, i.e., we want  to minimize the negative log-likelihood 

\begin{equation} 
\Lcal(\{J_k\},h|\mathbf{s}) = -\frac{1}{M} \sum_{\mu=1}^M \log P(\mathbf{s}^{\mu}) 
\label{logLikelihood}
\end{equation}
\noindent where $M$ is the number of samples at our disposal. The minimization of the above function is very difficult in general  due to the intractability of the partition function $Z$ in (\ref{eq:maxEntProb}), so we have to resort to an approximate method. Among many possible approximate methods for solving the above problem we chose the \emph{pseudo-likelihood} maximization approach, introduced in \cite{ravikumar2010high}, for reasons that will become clear below. The original paper treats models with binary variables such as the Ising model.  A generalization to multi-valued variables, such as in our case, can be found in \cite{ekeberg2013improved} where the authors use a very similar model to ours  to infer interactions between amino-acids in protein-protein interactions. 
 
In the pseudo-likelihood approximation one replaces, in the minimization task, the full probability of the model with the conditional probabilities of the variables given their neighbors and, in that way, replaces the original problem with a set of logistic regression problems. Each neighborhood is inferred independently and then the information is combined to get the full interaction graph. The authors in \cite{ravikumar2010high} have shown that this approximation works well under certain conditions. In physics terminology, as long as  the variables are not interacting too strongly, treating the neighborhoods independently gives fairly good results. We have found empirically that the parameters inferred from musical data fall into this category since  we are able to reproduce the corpus note frequencies fairly well. As we described in the previous section, the model can be decomposed in a number of identical copies of some basic module.  This particular structure makes the pseudo-likelihood a particularly appropriate method since the parameters on a neighbourhood have to be inferred only once.   Moreover, the above picture leads to a natural way to segment the corpus into samples: the samples used in the training phase are all the substrings of size $2 K_{\text{max}} +1$ in the corpus, each providing the necessary information to model  the way a variable depends on its local context, i.e., on $K_{\text{max}}$ variables to its left and to its right, see Fig.~\ref{fig:data}. There is a harmless redundancy in the data since most  note pairs are used twice when inferring the interaction potentials $J_k$, except for a negligible number of pairs of order $O(K_{\text{max}})$ at the corpus' borders.

In our case, the conditional probability used in the pseudo-likelihood approach is written
\begin{equation}
P(s_0^{(\mu)} | \mathbf{s}_{\setminus 0}^{(\mu)})  = 
  \frac{\exp\left\{ h(s_0^{(\mu)}) + \sum\limits_{l=1}^K \left(J_l(s_{-l}^{(\mu)},s_0^{(\mu)}) 
  + J_l(s_0^{(\mu)},s_l^{(\mu)}) \right) \right\}}
{\sum\limits_{{\s=1}}^{q} \exp\left\{ h(\s) + \sum\limits_{l=1}^K \left(J_l(s_{-l}^{(\mu)},\s) 
  + J_l(\s,s_l^{(\mu)}) \right) \right\}} \;\;\;,
\label{conditionalProb}
\end{equation}
where the index $\mu$ represents variables belonging to the $\mu^{\text{th}}$ sample,
$s_0^{(\mu)}$ is the central variable, $\mathbf{s}_{\setminus 0}^{(\mu)}$ represents
the remaining variables in the neighborhood, and $s_{-l}^{(\mu)}$ and $s_{l}^{(\mu)}$
the $l^{\text{th}}$ variable to the left and to the right of the central one respectively.  
The advantage of the above method lies in the tractability of the normalization
in (\ref{conditionalProb}) as opposed to the one in (\ref{eq:maxEntProb}).
The log-pseudo-likelihood function is
\begin{equation}
\Lcal_{\text{pseudo}}(\{J_k\},h|\Ssf) = -\frac{1}{M} \sum_{\mu=1}^M \log P(s_0^{(\mu)} | \mathbf{s}_{\setminus 0}^{(\mu)})
\;\;\;\;,
\label{loglikePseudoL}
\end{equation}
\noindent where the sum rums over all samples, i.e., all substrings of length $2 K_{\text{max}} +1$ of the  corpus. Minimizing the above function yields the potentials of the whole model since they are repeated on every neighborhood. In addition, one usually adds a regularization term to avoid over-fitting issues  (see Section S1 of the SI). For the details concerning the optimization procedure see Section S2 of the SI.

\appendix

\section*{S1 Regularization procedure}
\label{S1_Regularization}
   
In addition to the log-likelihood defined in the main text (reported here below for convenience) 

\begin{equation}\label{eq:loglikePseudoL}
\Lcal_{\text{pseudo}}(\{J_k\},h|\Ssf) = -\frac{1}{M} \sum\limits_{\mu=1}^M \log P(s_0^{(\mu)} | \mathbf{s}_{\setminus 0}^{(\mu)})
\;\;\;\;,
\end{equation}

\noindent one usually adds a regularization term. Regularization is motivated from multiple points of view, the main of which is to avoid over-fitting. From the point of view of our model, regularization has an additional benefit. It eliminates the degeneracy in the choice of parameters in the model. The model given by 

\begin{equation}\label{eq:hamiltonian}
\Hcal(s_1, \dots, s_N) = 
 -\sum\limits_{i=1}^N h(s_i) - \sum\limits_{k=1}^{K_{\text{max}}} \sum\limits_{\substack{i<j \\ |i-j|=k}} J_k(s_i,s_j) \;\;\;.
\end{equation}

\noindent exhibits \emph{gauge invariance}, i.e., there are different choices of parameters $J_k$ and $h$ which assign the same probabilities to the same variable configurations (see for example \cite{Weigt2009}). In order to have unique and reproducible results for a given training set we wish to remove this degeneracy. The addition of a $\ell_p$-norm regularizer does that by yielding the solution which minimizes the $\ell_p$-norm of the interaction matrices $J_k$.
 
Specifically, here we use $\ell_1$-norm regularization, pioneered in \cite{tibshirani1996regression}, and adapted in the context of sparse model selection in \cite{ravikumar2010high}. The benefits of $\ell_1$-norm regularization are twofold. First, it yields sparse results by forcing a substantial number of parameters to zero. This is reasonable in our case since our starting information is sparse: out of all the possible note-pairs only a fraction is actually used. Thus it wouldn't make much  sense to infer a number of parameters much larger than the number of independent quantities actually observed. The second advantage of the $\ell_1$-norm is that it conserves the convexity of the objective  function.

Concretely, instead of the log-likelihood function in eq.~(\ref{eq:loglikePseudoL}) we use:
\begin{equation}
\Lcal_{\text{pseudo, reg}}(\{J_k\},h|\Ssf) = -\frac{1}{M} \sum\limits_{\mu=1}^M 
\log P(s_0^{(\mu)} | \mathbf{s}_{\setminus 0}^{(\mu)}) + \frac{\lambda}{M} \sum\limits_{k=1}^{K_{\text{max}}} \|J_k\|_1 
\;\;\;\;.
\label{eq:pseudoLReg}
\end{equation}

\noindent By tuning the parameter $\lambda$ one can force the minimization procedure to set more or less values of the matrices $J_k$ equal to zero.  In practice, we found that values between $\lambda=1$ and $3$ yielded the smallest MSE between model and corpus pair-note probabilities.

\section*{S2 Optimization} 
\label{S2_Optimization}

We start by writing explicitly the negative log-likelihood. Since we use the pseudo-likelihood approach we will use the following conditional probability 

\begin{equation}\label{eq:conditionalProb}
P(s_0^{(\mu)} | \mathbf{s}_{\setminus 0}^{(\mu)})  = 
  \frac{\exp\left\{ h(s_0^{(\mu)}) + \sum\limits_{l=1}^K \left(J_l(s_{-l}^{(\mu)},s_0^{(\mu)}) 
  + J_l(s_0^{(\mu)},s_l^{(\mu)}) \right) \right\}}
{\sum\limits_{\s=1}^q \exp\left\{ h(\s) + \sum\limits_{l=1}^K \left(J_l(s_{-l}^{(\mu)},\s) 
  + J_l(\s,s_l^{(\mu)}) \right) \right\}} \;\;\;,
\end{equation}

\noindent The negative logarithm of the pseudo-likelihood together with the regularization term  
then reads
\begin{equation}\label{eq:pseudoLogLikeReg}
\begin{split}
\Lcal_{\text{pseudo, reg}}(\{J_k\},h|\Ssf) = -\frac{1}{M} \sum\limits_{\mu=1}^M &\log P(s_0^{(\mu)} | \mathbf{s}_{\setminus 0}^{(\mu)})  \\
= -\frac{1}{M} \sum\limits_{\mu=1}^M &\Bigg\{ 
h(s_0^{(\mu)}) + 
\sum\limits_{k=1}^{K_{\text{max}}} 
\left( J_k(s_{-l}^{(\mu)},s_0^{(\mu)}) + J_k(s_0^{(\mu)},s_{+l}^{(\mu)}) \right) 
 \\
&  \phantom{blablablablablabla}- \log Z^{(\mu)}\Bigg\} + \frac{\lambda}{M}\sum\limits_{k=1}^{K_{\text{max}}} \|J_k\|_1
 \;\;\;\text{,}
\end{split}
\end{equation}

\noindent where the partition function is 

\begin{equation}\label{eq:partitionFunction}
Z^{(\mu)} = \sum\limits_{\s=1}^q \exp\left\{ h(\s) + \sum\limits_{l=1}^{K_{\text{max}}} \left(J_l(s_{-l}^{(\mu)},\s) 
  + J_l(\s,s_l^{(\mu)}) \right) \right\} \;\;\;.
\end{equation}

\noindent For the minimization of the above function we used an $\ell_1$-regularized problem solver written in Matlab by Mark Schmidt \cite{schmidt2010graphical,schmidt2007fast}. Specifically, we used the projected scaled sub-gradient (Gafni-Bertsekas variant). The solver makes use of the gradient of the above function, provided by the user. The gradient elements have two forms depending on whether one is differentiating with respect to a local field or an interaction potential. Omitting regularization terms
for simpliity, the two forms are respectively

\begin{equation}\label{eq:gradLocField}
\begin{split}
\frac{\partial \Lcal}{\partial h(r)} &= - \frac{1}{M}\sum\limits_{\mu=1}^M \left(   \dd(s_0^{(\mu)},r) - 
\frac{\partial}{\partial h(r)} \log Z^{(\mu)} \right) \\
%% &=  - \frac{1}{M}\sum\limits_{\mu=1}^M \Bigg(   \dd(s_0^{(\mu)},r) \\
%% &\phantom{blablabla} -\frac{1}{Z^{(\mu)}} \sum\limits_{\s=1}^q \dd(r,\s) \exp \left\{ h(\s) + \sum\limits_{l=1}^{K_{\text{max}}} \left(J_l(s_{-l}^{(\mu)},\s) 
%%   + J_l(\s,s_l^{(\mu)}) \right) \right\} \Bigg) \\
&= - \frac{1}{M}\sum\limits_{\mu=1}^M \Bigg(   \dd(s_0^{(\mu)},r) 
- \frac{1}{Z^{(\mu)}} \exp \left\{ h(r) + \sum\limits_{l=1}^{K_{\text{max}}} \left(J_l(s_{-l}^{(\mu)},r) 
  + J_l(r,s_l^{(\mu)}) \right) \right\} \Bigg) \\
&
\end{split}
\end{equation}

\noindent and

\begin{equation}\label{eq:gradCoupling}
\begin{split}
\frac{\partial \Lcal}{\partial J_k(r,r')} &= - \frac{1}{M}\sum\limits_{\mu=1}^M \left(   \dd(s_{-k}^{(\mu)},r)\dd(s_{0}^{(\mu)},r') 
+\dd(s_{0}^{(\mu)},r)\dd(s_{+k}^{(\mu)},r')
- \frac{\partial}{\partial J_k(r,r')} \log Z^{(\mu)} \right)  \\
&=- \frac{1}{M}\sum\limits_{\mu=1}^M \Bigg(   \dd(s_{-k}^{(\mu)},r)\dd(s_{0}^{(\mu)},r') +\dd(s_{0}^{(\mu)},r)\dd(s_{+k}^{(\mu)},r')\\
&\phantom{blablabla} - \frac{1}{Z^{(\mu)}} \dd(s_{-k}^{(\mu)},r) \exp \left\{ h(r') + \sum\limits_{l=1}^{K_{\text{max}}} \left(J_l(s_{-l}^{(\mu)},r') 
  + J_l(r',s_l^{(\mu)}) \right) \right\}  \\
&\phantom{blablabla} - \frac{1}{Z^{(\mu)}} \dd(s_{+k}^{(\mu)},r') \exp \left\{ h(r) + \sum\limits_{l=1}^{K_{\text{max}}} \left(J_l(s_{-l}^{(\mu)},r) 
  + J_l(r,s_l^{(\mu)}) \right) \right\} \Bigg) \;\;\;\;. \\
&
\end{split}
\end{equation}

\noindent Note that the first terms in the above equation are counting the number of occurrences of some event in the samples. The first term in eq.~(\ref{eq:gradLocField}) counts the number of samples where the central variable takes the value $r$. The first and second terms in eq.~(\ref{eq:gradCoupling}) count the number of  samples where the pair $r,r'$ appears at distance $k$ and where one of the two variables is the central one. These quantities can be precomputed. The remaining terms (except the regularization) are the model predictions for these same quantities, so finding the minimum of the function in eq.~(\ref{eq:pseudoLogLikeReg}) equates the model frequencies with the empirical ones.

\section*{S3 Musical Corpora}
\label{subsec:corpora}

We used a variety of musical corpora in our experiments. Since our model is by construction 
monophonic we either used monophonic music or extracted monophonic sequences from polyphonic 
pieces. Additionally, we ignored rhythm and discarded all information about note onsets and 
durrations.
All corpora where taken from the following two databases:
\begin{itemize}
\item The \textsf{\textbf{Weimar Jazz Database (WJDB)}}\cite{WeimarJazzDatabase} contains detailed
transcriptions of famous jazz improvisations. As of March 2015 the database contains 257 songs.
The majority of improvisations come from brass and wind instruments so they are monophonic by 
default.
\item \textsf{\textbf{Kunst der Fuge (KdF)}}\cite{KdF} is a classical music MIDI files database. It
contains thousands of classical music pieces in MIDI format. Most of the tunes are polyphonic
so, in that case, we extracted separate tracks from the MIDI files when possible and chose
our sequences among them. When necessary, we discarded simultaneous notes by keeping the highest 
one.
\end{itemize}

To be more specific, the sequences  used as training corpora in this paper were exctracted from 
the following works.

\begin{center}
  \begin{table}[h!]
    \label{SItab:corpora}
  \caption{Musical pieces used as training corpora in this work}
    \begin{tabular}{ | c | c | p{5cm} |}
    \hline
    \bf{Figure} & \bf{Data Base} & \bf{Work} \\ \hline
    Fig.~3 & WJDB & The whole database \\ \hline
    \begin{tabular}{c}
    Fig.~4, Fig.~5, Fig.~6 \\
    Fig.~\ref{SIfig:LCS},Fig.~\ref{SIfig:InovPatterns}  
    \end{tabular}
    & KdF & Violin Partita No.1 in B minor, BWV 1002 (Bach, Johann Sebastian) \\ \hline
    Fig.~\ref{SIfig:Innovation_vs_similarity_beethoven} & KdF & 
    \begin{itemize}
      \item 11 Bagatelles, Op.119 (Beethoven, Ludwig van)
      \item Movement I, Symphony No.8, Op.93 (Beethoven, Ludwig van)
      \item 3 Marches, Op.45 (Beethoven, Ludwig van)
    \end{itemize}
    \\ \hline
    Fig.~\ref{SIfig:Innovation_vs_similarity_SCL} & KdF & 
    \begin{itemize}
      \item Andante und Variationen, Op.46 (Schumann, Robert)
      \item Mazurka No.1, Op.56 (Chopin, Frédéric)
      \item Liebestr\"aume No.3, S.541 (Liszt, Franz)
    \end{itemize}
    \\ \hline
    Fig.~\ref{SIfig:Innovation_vs_similarity_LCS}
    & KdF & Violin Partita No.1 in B minor, BWV 1002 (Bach, Johann Sebastian) \\ \hline
    \end{tabular}
    \end{table}
\end{center}

In addition, in the figures Fig.~6, 
Fig.~\ref{SIfig:Innovation_vs_similarity_beethoven}, 
Fig.~\ref{SIfig:Innovation_vs_similarity_SCL} and
Fig.~\ref{SIfig:Innovation_vs_similarity_LCS}, the borrowing and similarity features 
where also computed against a large number of sequences called \emph{AllClass}, extracted from the 
KdF database and coming from composers Albeniz, Bach, Beethoven, Chopin, Liszt and Schumann.
In these figures, as a reference for the evaluation of the models, we compute additionally
the borrowing and similarity features between the training corpus and the above mentioned 
sequences. We discriminate between sequences from the same composer (blue non-filled circles)
and sequences of all other composers (grey non-filled circles).
In order to avoid different tonalities 
issues we used sequences of intervals (see main text, section \textbf{borrowing and innovation}).

\section*{S4 Similarity vs. Innovation}
\label{subsec:inovation}

\noindent In the main text we have seen that our model is able to reproduce melodic patterns from the corpus of different sizes, although the model only explicitly enforces pairwise constraints. We have also quantified the interplay between borrowing and innovation in the artificial sequences generated starting from a given corpus. Here innovation means that not all melodic patterns in the generated sequences are identical to ones found in the corpus. Although these new patterns might not be part of the corpus, they are far from being random. They emerge from the effort to satisfy multiple competing pairwise constraints and for that reason they are ``musically'' interesting.   When a musician decides to substitute a few in a given phrase, the new notes will bear relations with the remaining notes that are probably found elsewhere in the corpus. This is exactly what our model does.

\begin{figure}[h!]
    \centering
    \includegraphics[width=0.9\linewidth]{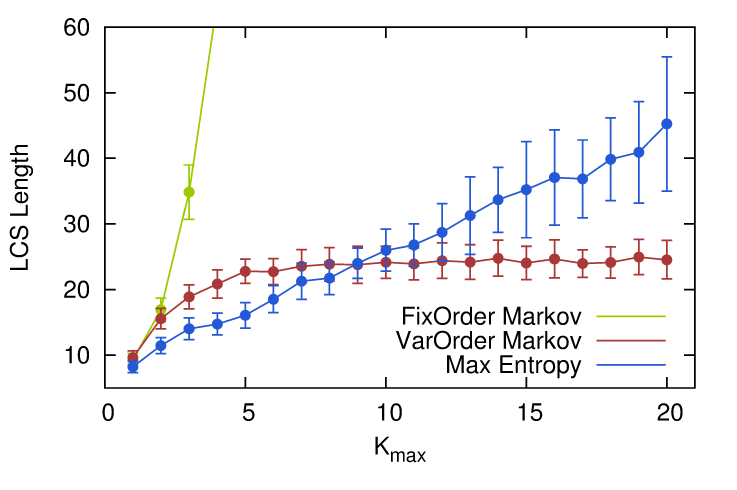}              
    \caption{{\bf Longest Common Substrings} between the original corpus (see Section S3 for details) and the sequences  (of length $N=5000$) artificially generate with the three models considered in  the main text. Results are obtained by averaging over $100$ sequences for each model.}
    \label{SIfig:LCS}
\end{figure}

We first focus on the level of borrowing or imitation. We study in particular the behavior of the \emph{Longest Common Substring} (LCS) between the original and generated sequences. This feature provides some insight about the balance between style imitation on one hand and plagiarism on the other. As we said earlier, style imitation consists, among other things, in re-arranging existing melodic patterns. If these patterns are too long the new sequence will give the impression of copying the corpus instead of imitating it.  If on the other hand  these patterns are too short, the style of the corpus will not be recognizable. It would be then desirable to be able to control the size of the longest common substring (LCS) in the generated  sequence in order to find the right balance between imitation and innovation.

Fig.~\ref{SIfig:LCS} illustrates how the LCS for fixed-order Markov grows very fast with $K_{\text{max}}$, here the order of the model, leading to total plagiarism. On the other hand for variable-order Markov models the LCS quickly saturates to a particular value. Beyond this point changing $K_{\text{max}}$ does not have any effects since a much small $k$ is always selected.  In contrast, in our Maximum Entropy  model the size of LCS scales linearly with $K_{\text{max}}$.  As we discussed earlier, this property is desirable since it allows to fine-tune the balance between style imitation and plagiarism.

\begin{figure}[h]
    \centering
    \includegraphics[width=0.9\linewidth]{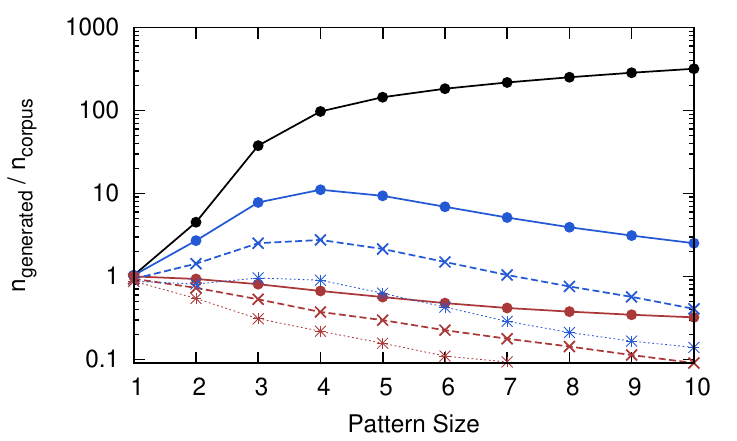}              
    \caption{{\bf Style imitation and innovation: patterns at Hamming distance $d=0$ and $d=1$ from the corpus patterns.} For a given pattern size (x-axis), the plot shows $n_{\text{generated}}^{(d)} / n_{\text{corpus}}$ for $d=0$ (red lines) and $d =1$ (blue lines) as a function of pattern size. The generated sequences are of size $N=500$ (dotted), $N=5000$ (dashed) and $N=50000$ (solid). As a reference we also counted the total number of patterns at $d=1$ from patterns in the corpus (black line).
For details on the corpus see Section S3}
    \label{SIfig:InovPatterns}
\end{figure}

In order to further picture the rates of borrowing and innovation of our model we did the following additional experiment. First we count the number of all distinct patterns that appear in the corpus as a function of pattern size $n_{\text{corpus}}(l)$. Then we generate sequences of various sizes and count the number of distinct patterns in them $n_{\text{generated}}^{(d)}(l)$ at a given Hamming distance $d$ from patterns also appearing in the corpus, or less. The Hamming distance between two patterns is simply the number of positions at which the sequences have different symbols. For example, if pattern $abcd$ appears in the corpus, then e.g., pattern $abxd$ with any $x$ is at distance $d =1$ if $x\ne c$ and $d=0$ for $x=c$. Finally we plot the fraction $n_{\text{generated}}^{(d)}(l) / n_{\text{corpus}}(l)$ for $d=0$ and $d =1$. When we count the number of patterns at $d= 0$ we are quantifying the degree of imitation of our model, since these patterns appear as such in the corpus. On the other hand, when we count  the number of patterns at $d=1$ we have some indication about the degree of innovation of our model. Fig.~\ref{SIfig:InovPatterns} reports the results. The red curves correspond to distance $d=0$ whereas the blue curves correspond to $d=1$. As a refference we have also counted the number of all possible patterns at distance $d=1$ from the patterns of the corpus, which we obtained by simple enumeration, and included the corresponding  curve (black line).

The figure can be read as follows. For a given size $N$ of the generated sequence, the region bellow the red line shows the rate of imitation, the region between the red and blue lines shows the rate of innovation and the region between the blue and black lines shows the amount of patterns, at distance $d=1$, that have been avoided by our model, because they failed to emerge from the competing pairwise constraints. This last point is important since it shows that the model is highly selective when generating new patterns.  We have listened to hundreds of patterns from the regions between the red and blue lines and found that the great majority of them where musically sensible alternatives to  patterns from the corpus. On the other hand, patterns from the region between the blue and black lines tend to sound ``wrong'' since they often violate multiple pairwise  constraints.

\section*{S5 Data-compression approach to measure cross-complexities and cross-entropies}
\label{subsec:zippers}

\begin{figure}[h!]
\centerline{\includegraphics[width=0.35\textwidth]{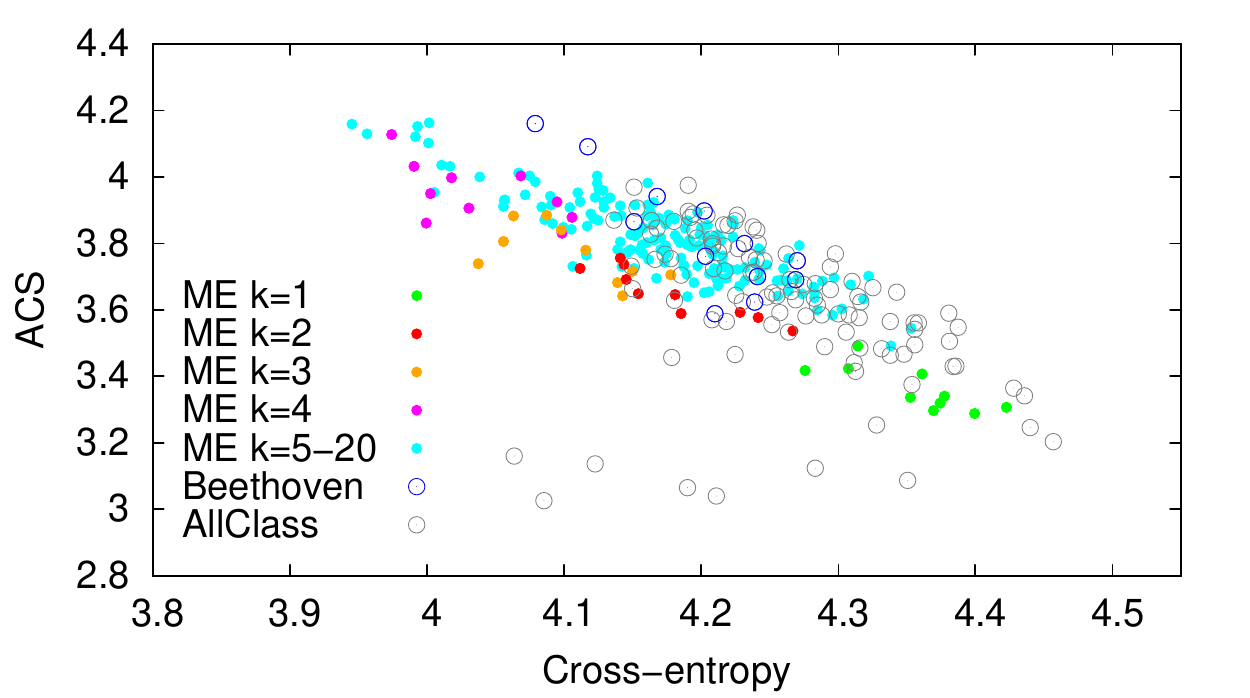}\includegraphics[width=0.35\textwidth]{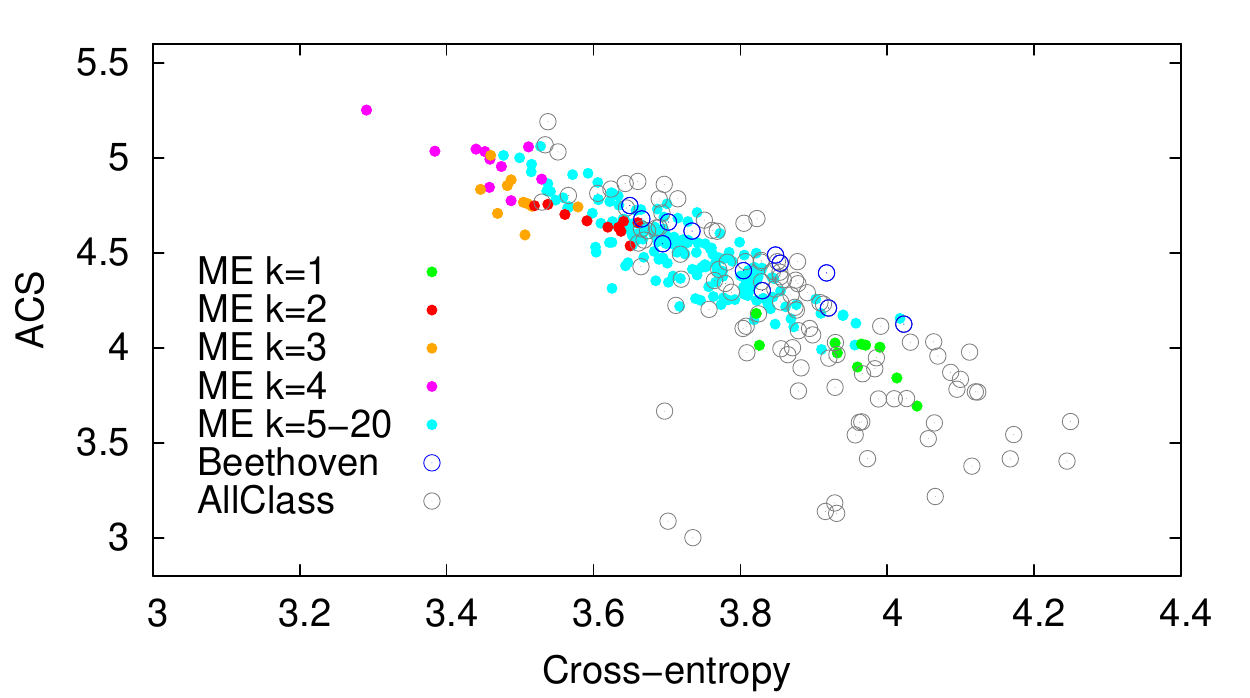}  \includegraphics[width=0.35\textwidth]{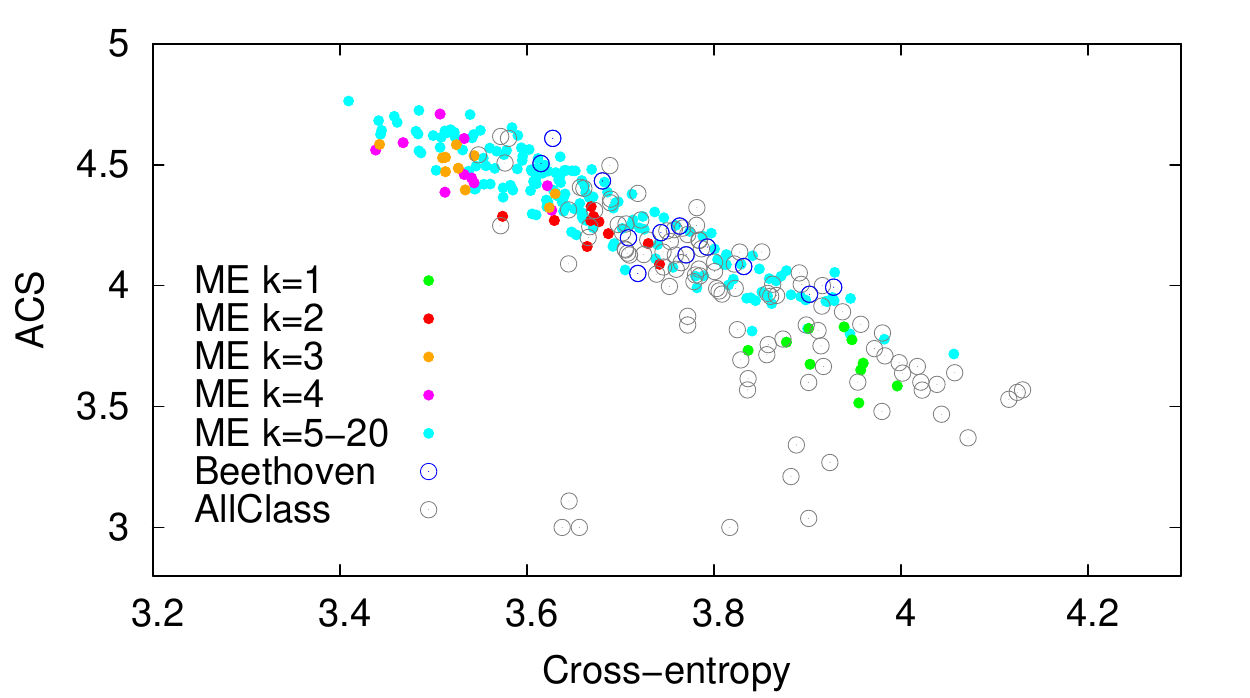}     }           
    \centerline{ \includegraphics[width=0.35\textwidth]{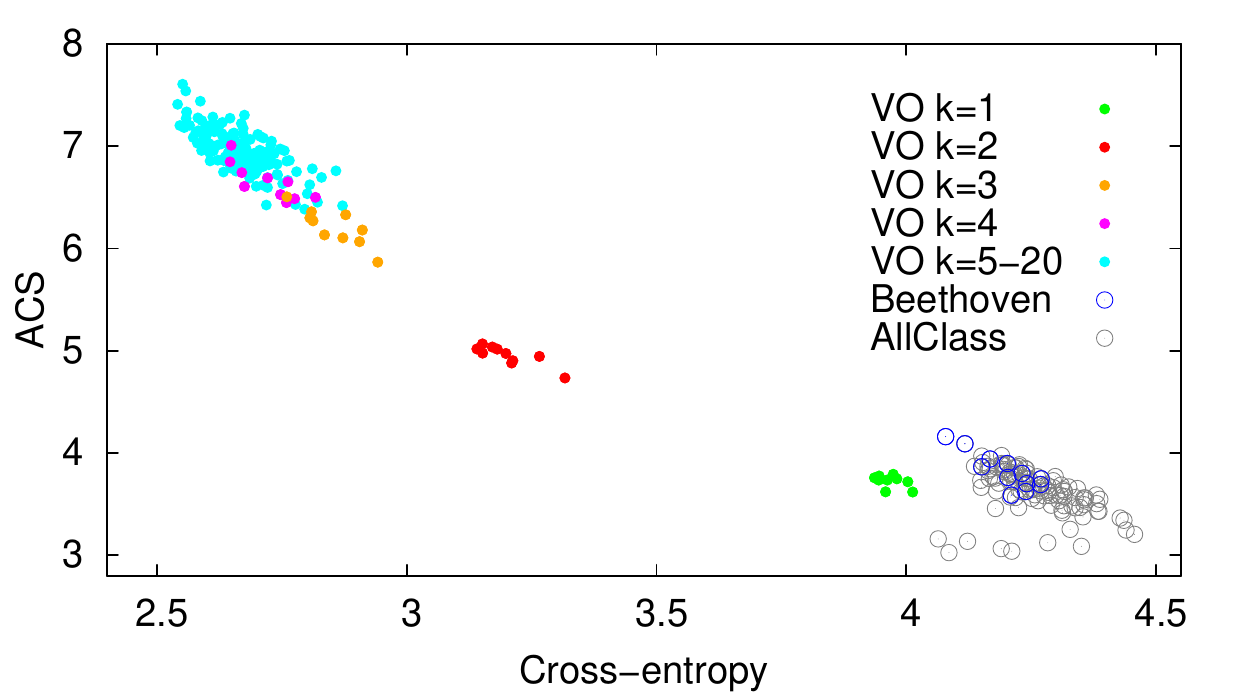}   \includegraphics[width=0.35\textwidth]{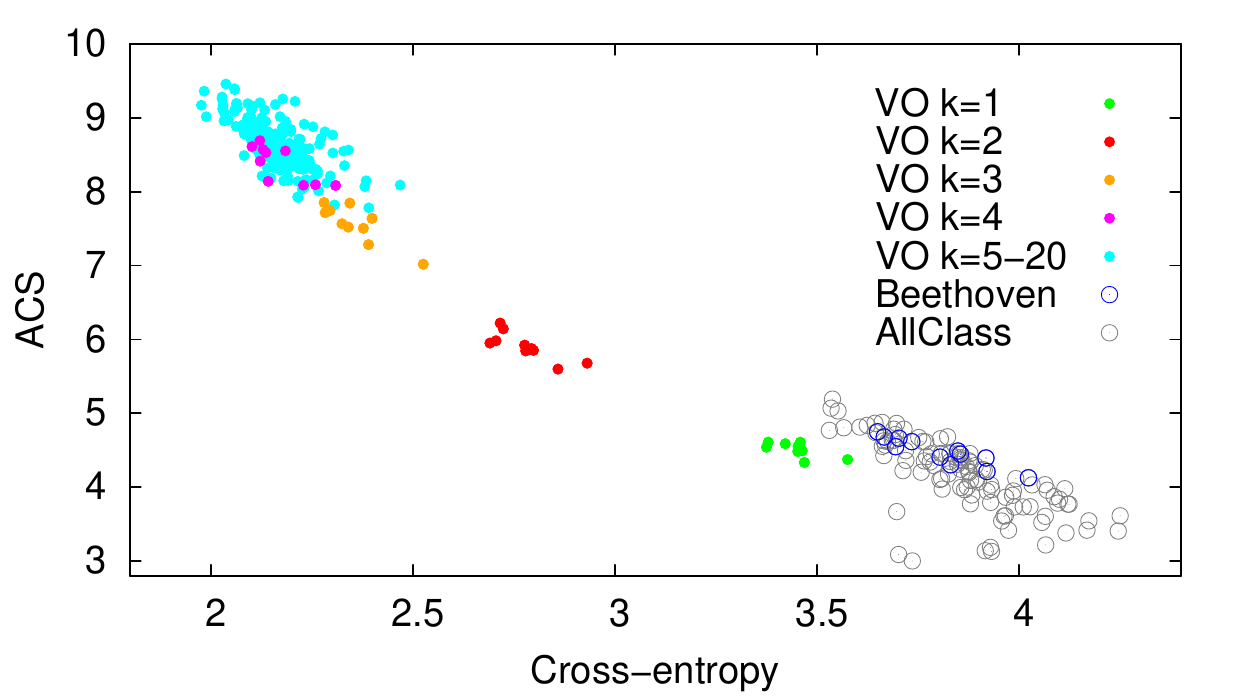}   \includegraphics[width=0.35\textwidth]{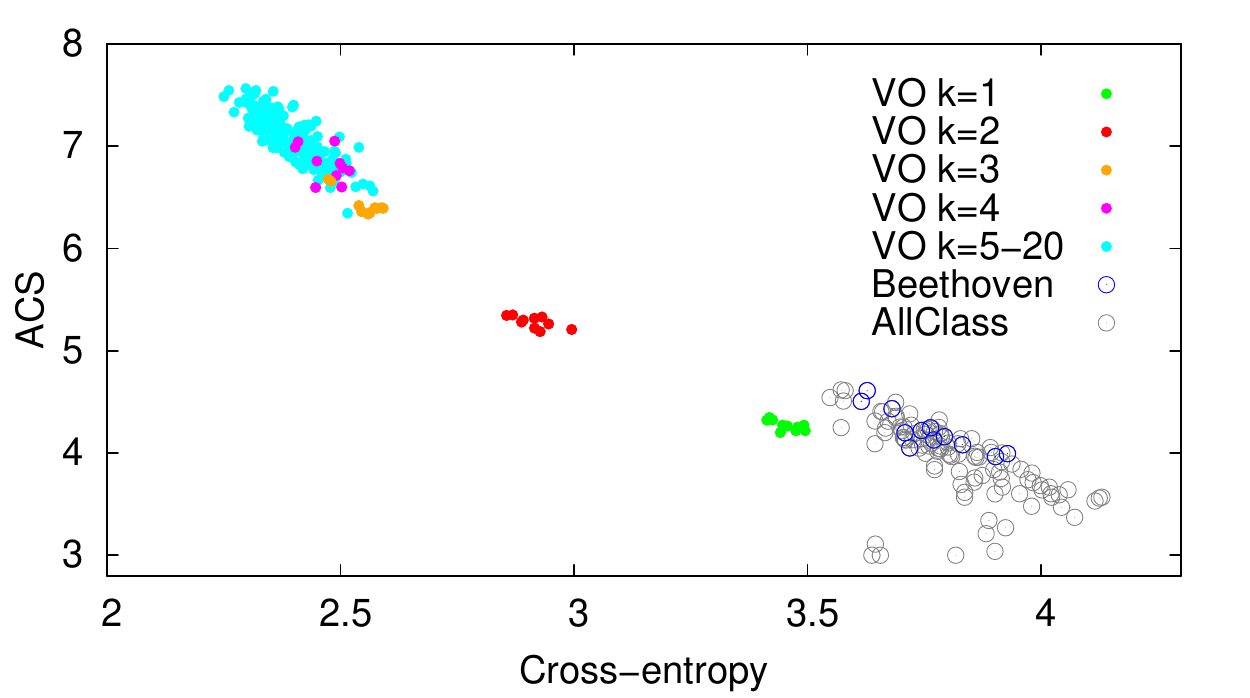}   }
   \centerline{   \includegraphics[width=0.35\textwidth]{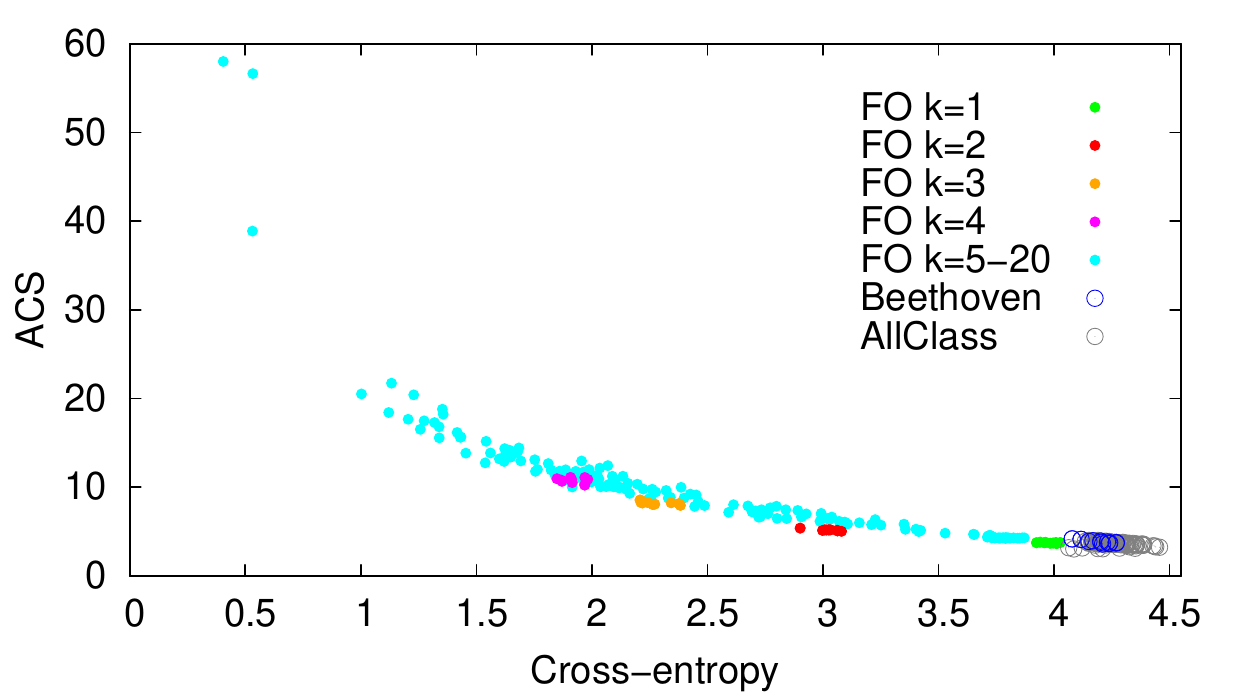}      \includegraphics[width=0.35\textwidth]{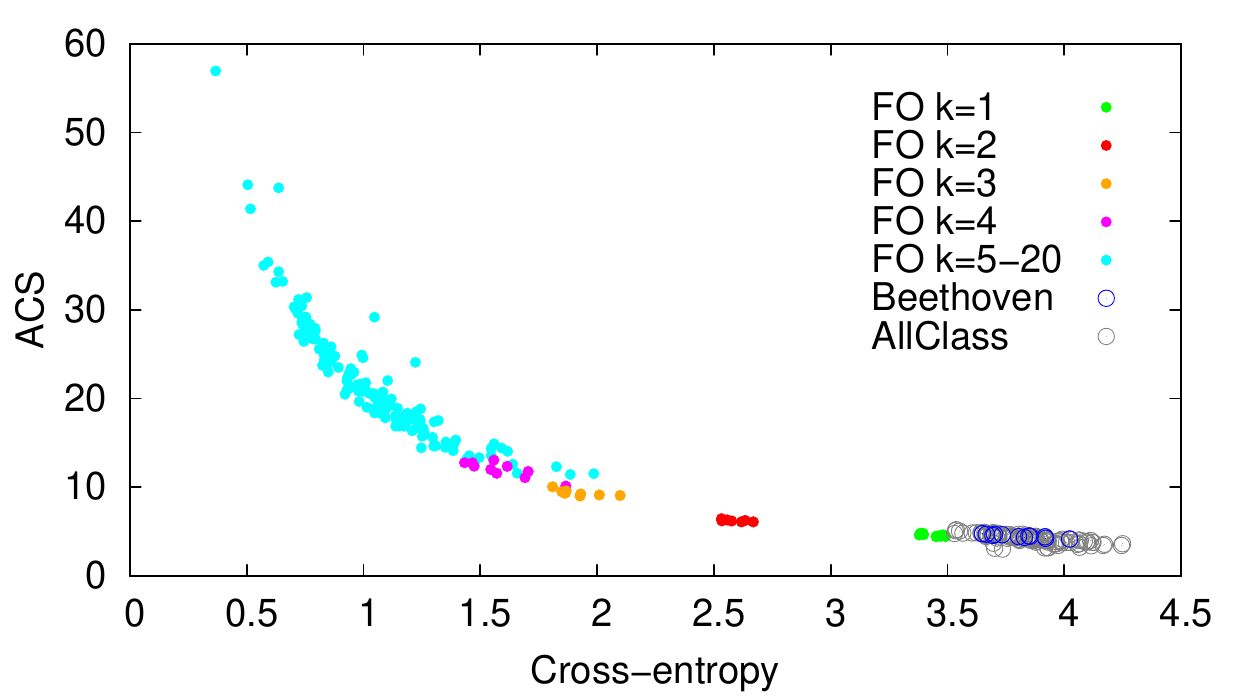}  \includegraphics[width=0.35\textwidth]{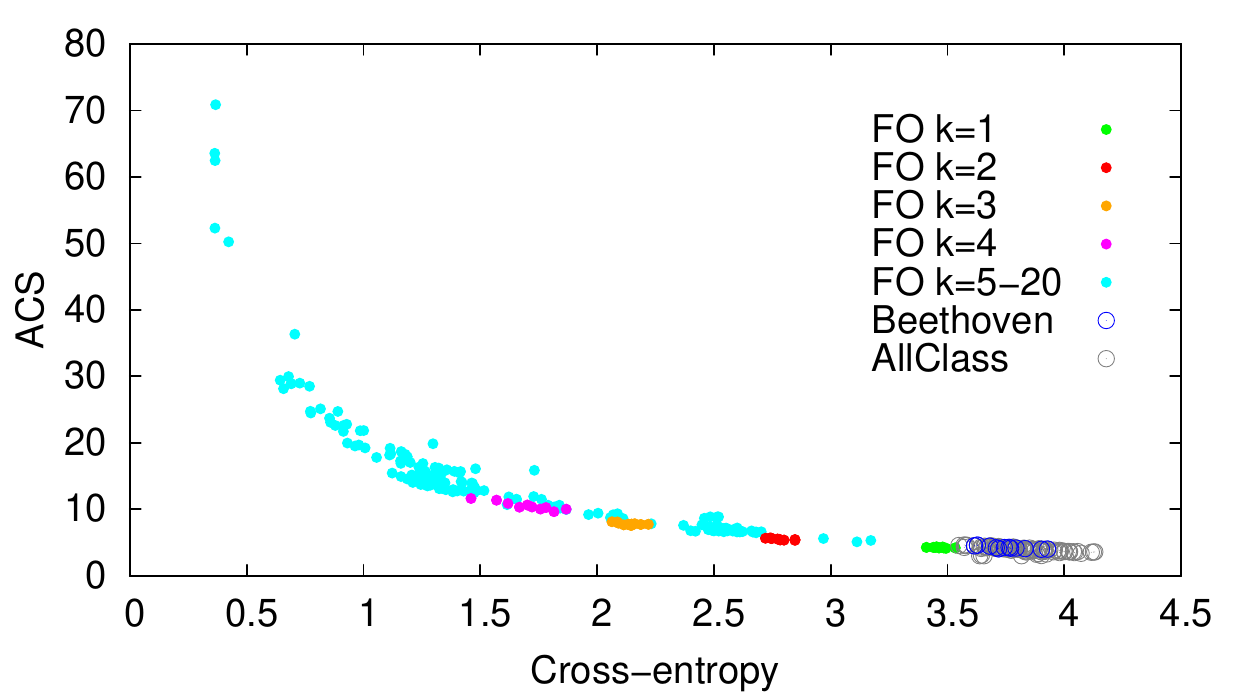}  }                   
    \caption{{\bf Borrowing vs. similarity in Beethoven.} This figure reports the Average Common Substring (ACS) vs. the values of the cross-entropies for all the artificial sequences generated with the Maximum Entropy (ME) model (top), the variable-order (VO) Markov model (middle) and the fixed-order (FO) Markov model (bottom). Everything is computed with respect to three sequences by Ludwig van Beethoven (see Section S3 for details). As in the main text, filled circles correspond to the artificial sequences. Colors code for the values of $K_{\text{max}}$ in each different model. In addition in each panel  the empty circles reports the same quantities for  other original sequences of Beethoven (represented with blue circles) and other classical authors - Bach, Schumann, Chopin, Liszt and Albeniz - (AllClass represented with grey circles).}
    \label{SIfig:Innovation_vs_similarity_beethoven}
\end{figure}

In this section we report some details about the data-compression techniques we used to estimate the similarity between pairs of sequences. We followed in particular the approach proposed in~\cite{bcl_2002,zippers_2005}, based on the Lempel-Ziv algorithm~\cite{LZ77}. We adopt in particular the notion of cross-complexity (from now onwards referred as cross-entropy) between two sequences of characters. Strictly speaking one refers to cross-entropy in information theory having in mind two sequences emitted from two specific sources. In Algorithmic Complexity Theory one deals with sequences without any reference to the sources that emitted them. In this case one speaks of cross-complexity between two sequences~ \cite{cerra_2012}.

\begin{figure}[h!]
\centerline{\includegraphics[width=0.35\textwidth]{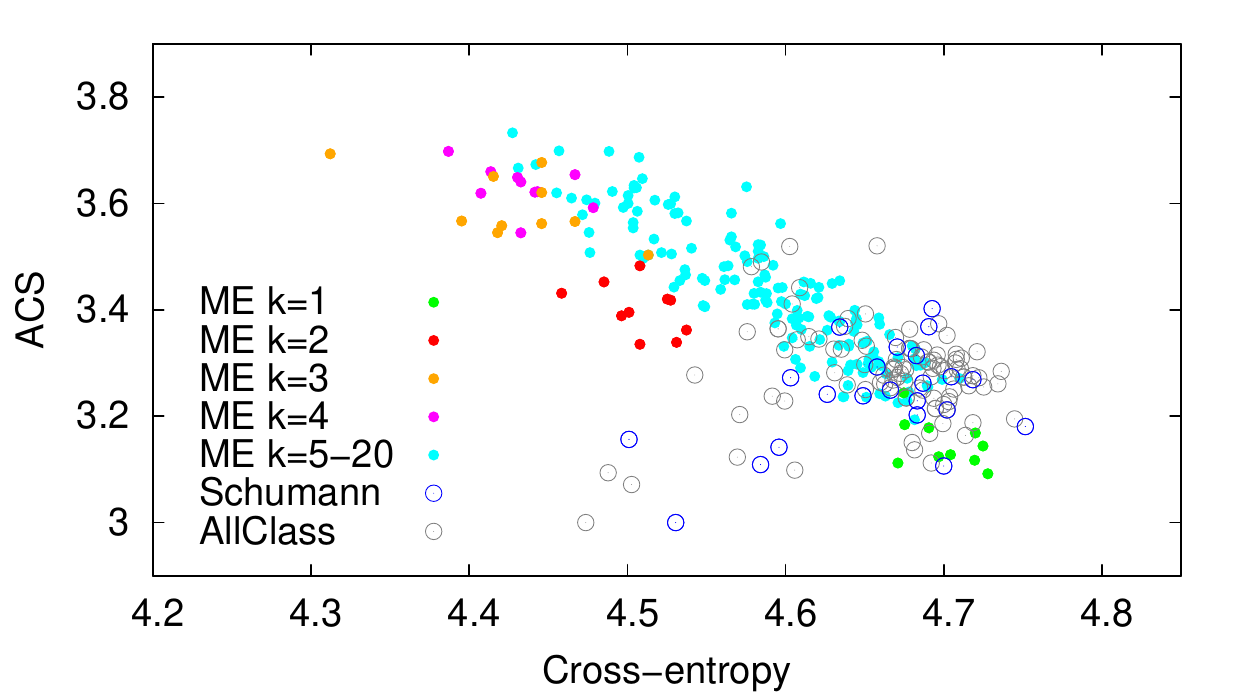}\includegraphics[width=0.35\textwidth]{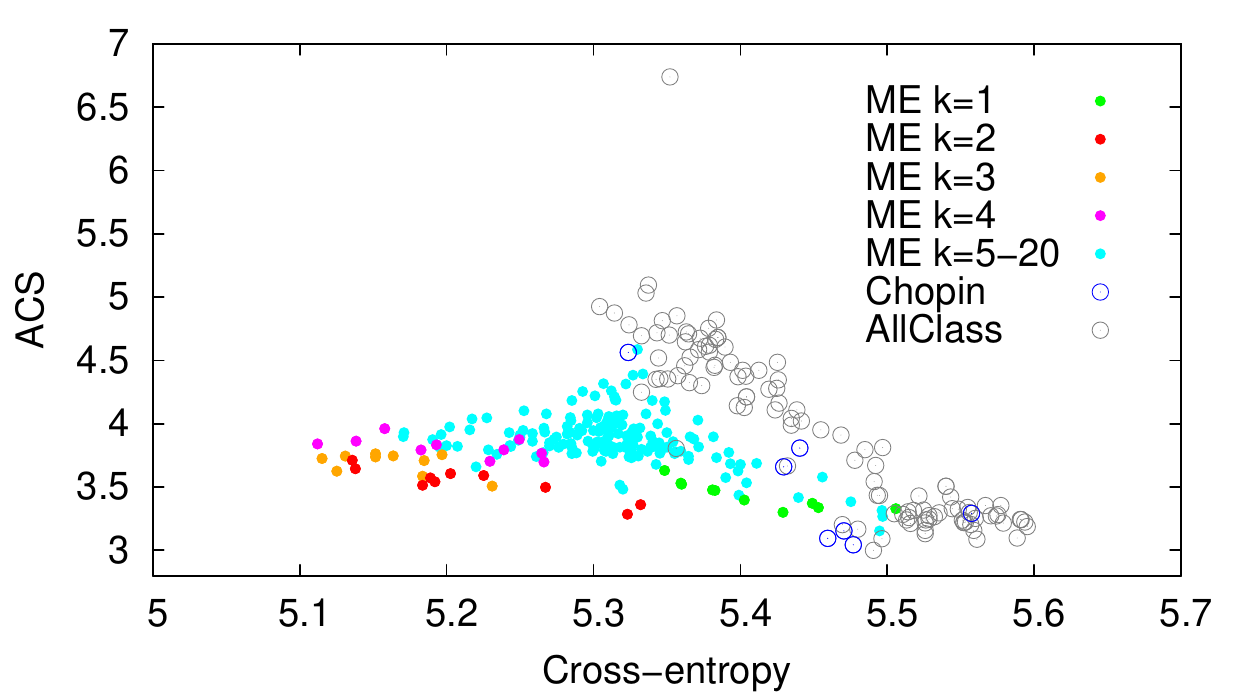}  \includegraphics[width=0.35\textwidth]{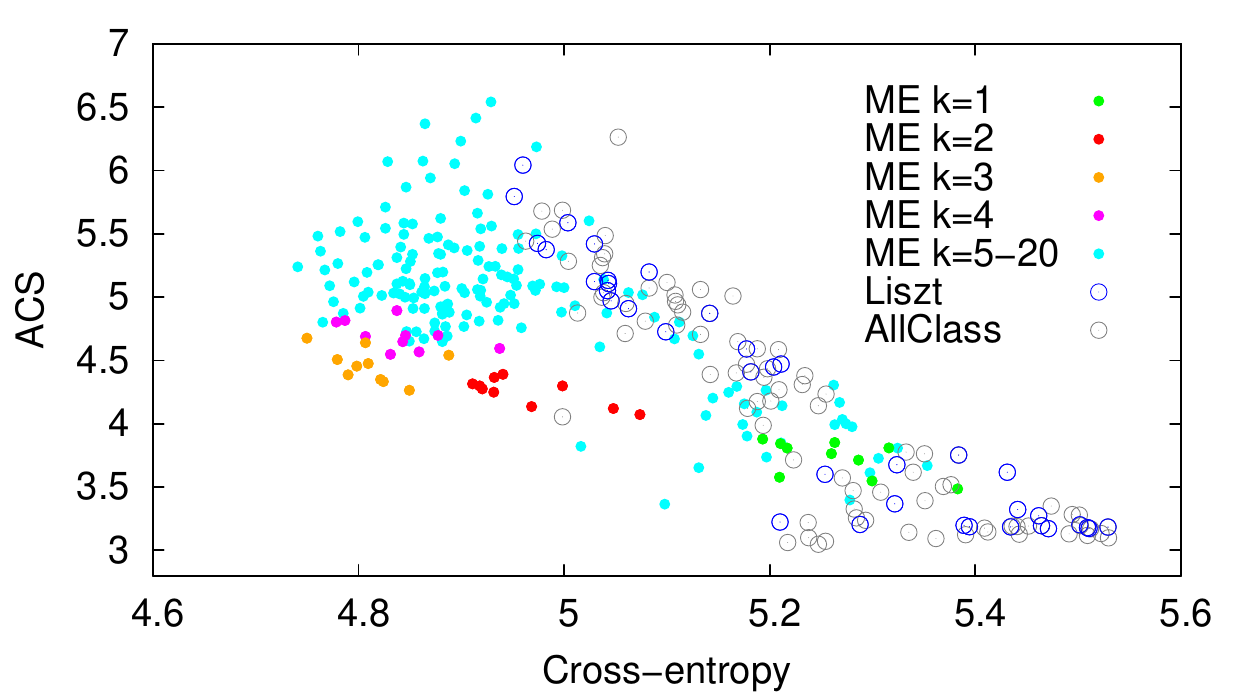}     }           
    \centerline{ \includegraphics[width=0.35\textwidth]{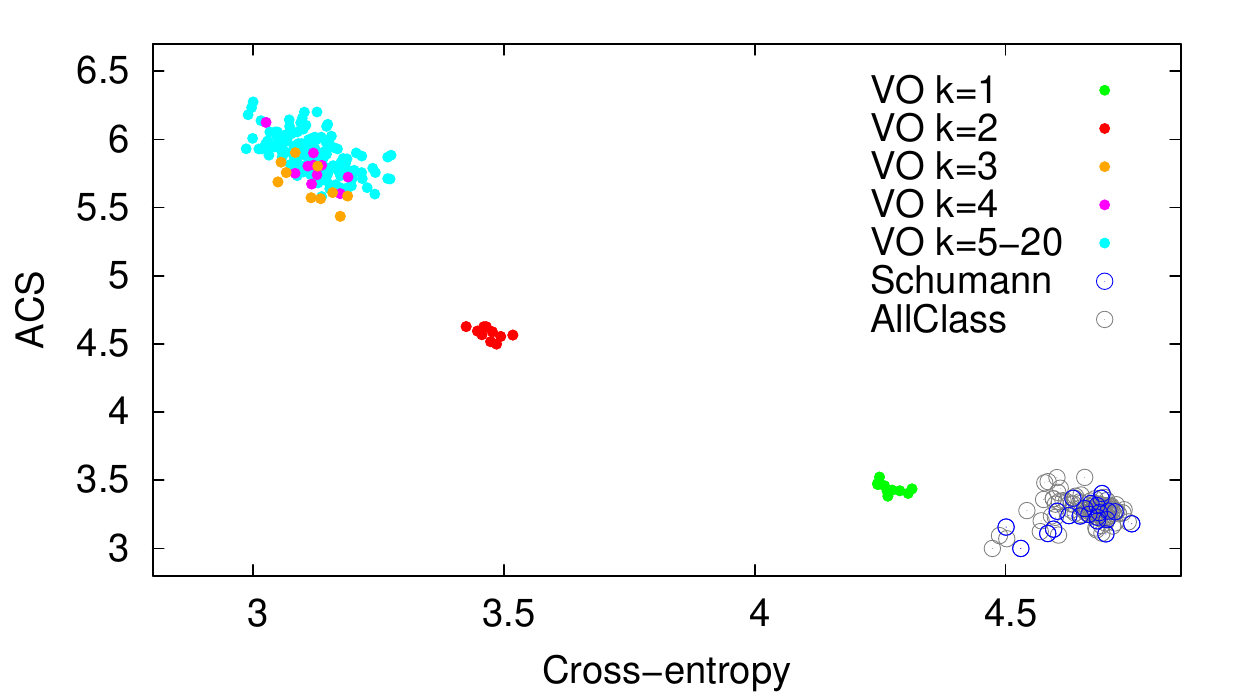}   \includegraphics[width=0.35\textwidth]{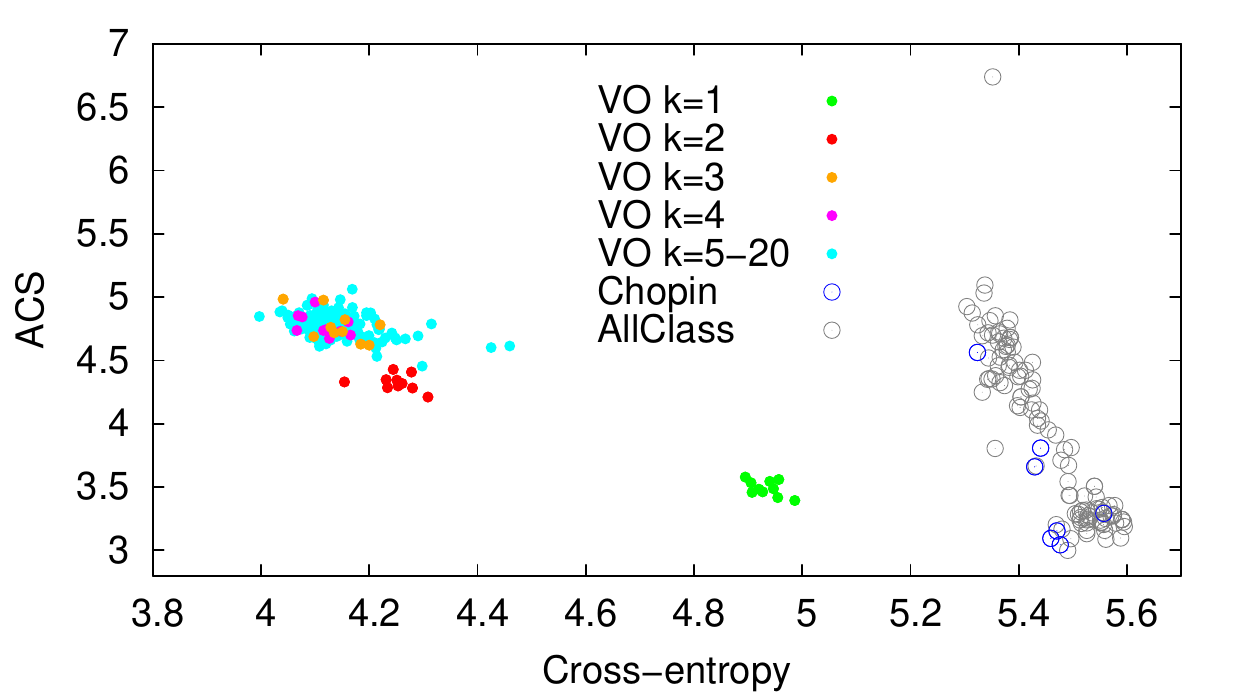}   \includegraphics[width=0.35\textwidth]{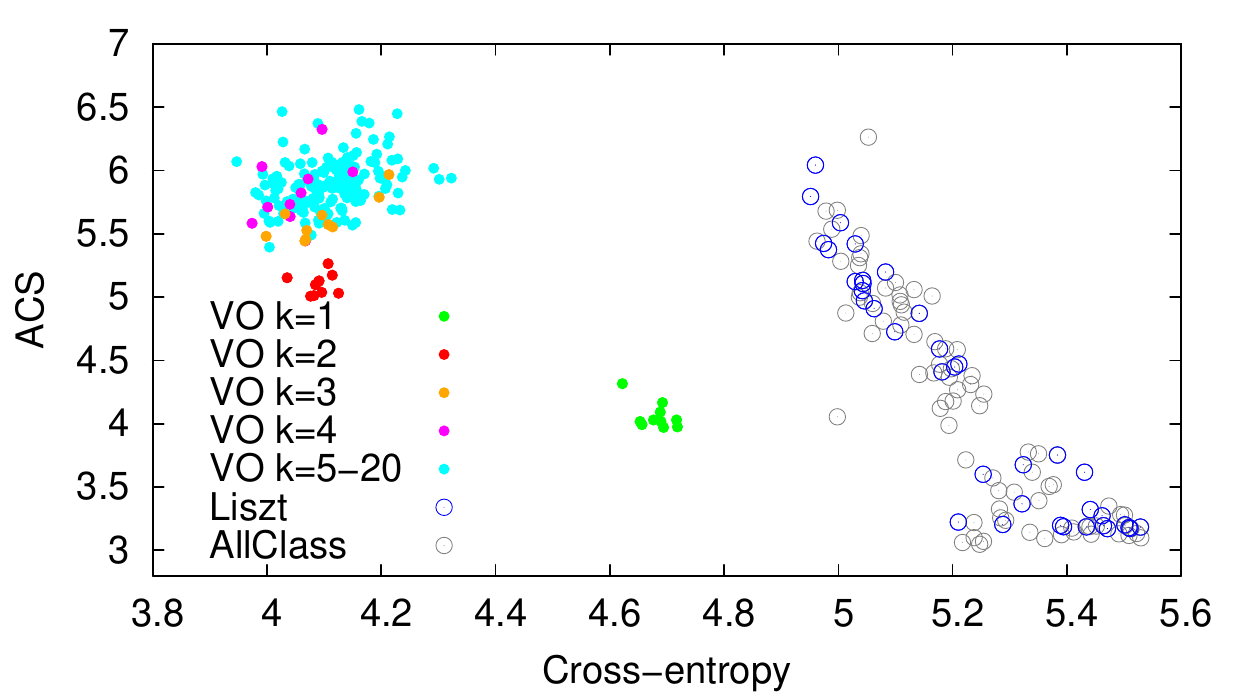}   }
   \centerline{   \includegraphics[width=0.35\textwidth]{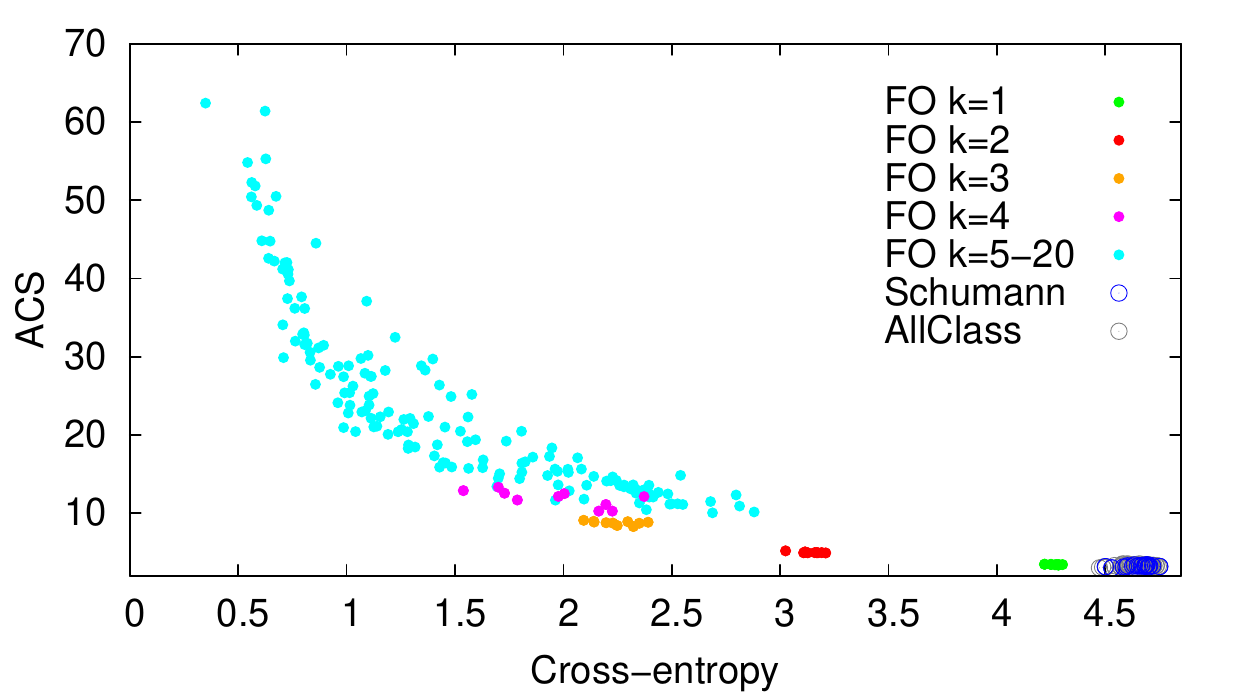}      \includegraphics[width=0.35\textwidth]{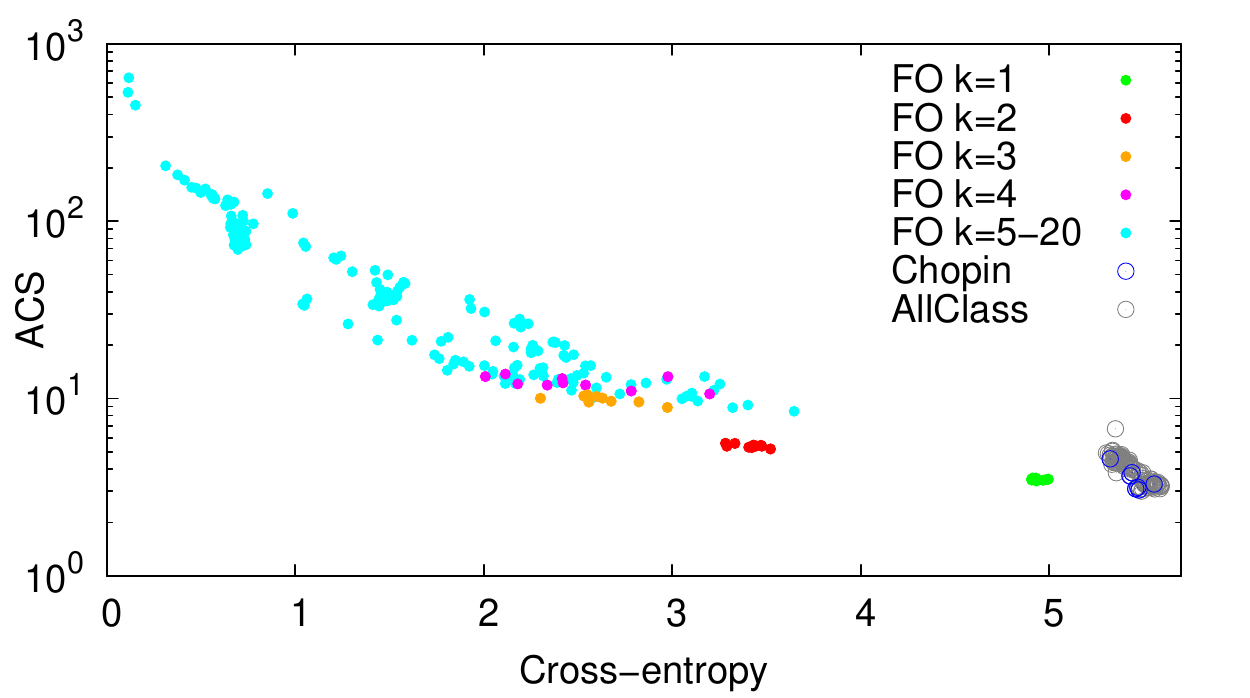}  \includegraphics[width=0.35\textwidth]{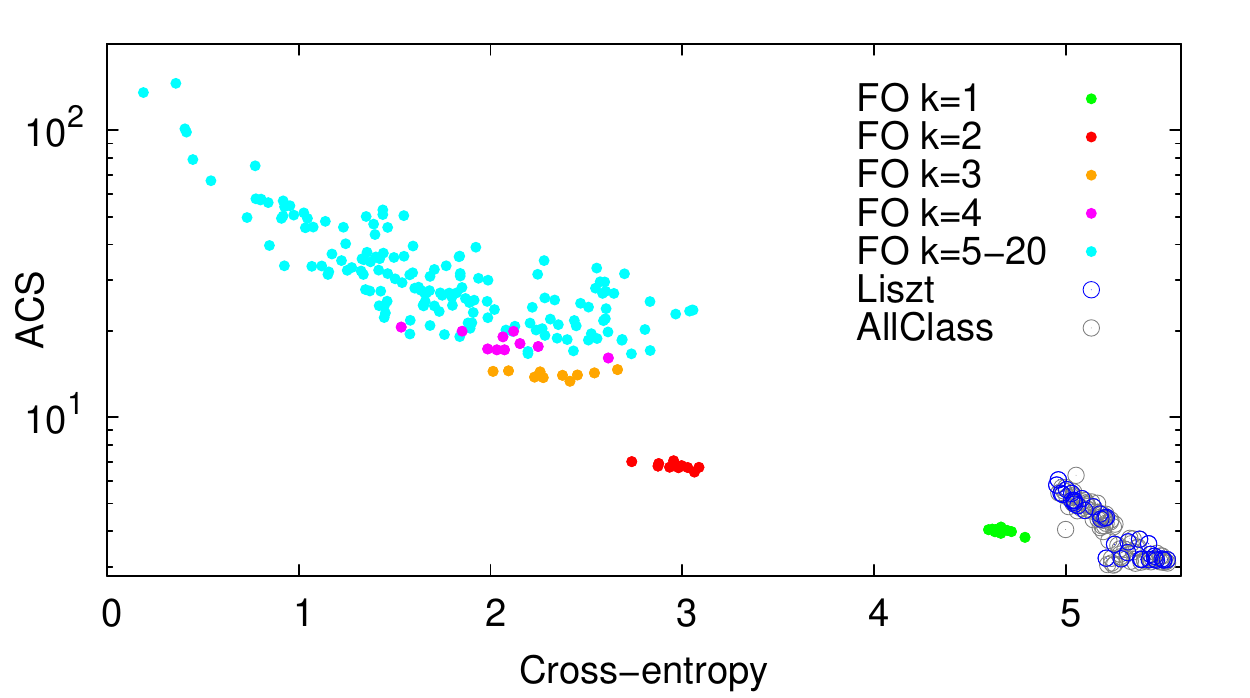}  }                   
    \caption{{\bf Borrowing vs. similarity in Schumann, Chopin and Liszt.} This figure reports the same results as in Fig.~\ref{SIfig:Innovation_vs_similarity_beethoven} for works by Robert Schumann (left column), Fr\'ed\'eric Chopin (center column) and Franz Liszt (right column). For each composer AllClass means the set of all the composers (Bach, Beethoven, Schumann, Chopin, Liszt and Albeniz) excluding the composer considered. For details on the corpora used see Section S3.}
    \label{SIfig:Innovation_vs_similarity_SCL}
\end{figure}

The cross-entropy between two sequences $\cal{A}$ and $\cal{B}$ is defined as the number of bits needed to encode each character emitted by the sequence $\cal{B}$ with the optimal coding for $\cal{A}$.  Consider for instance two ergodic sources A and B emitting sequences of zeroes and ones: A emits $0$ with probability $p$ and $1$ with probability $1-p$ whereas B emits $0$ with probability $q$ and $1$ with probability $1-q$. The previously described compression algorithm can encode a sequence emitted by A almost optimal-coding a $0$ with $-\log_2 p$ bits and a $1$ with $-\log_2 (1-p)$ bits. However, this A-optimal coding is not optimal for the sequence emitted by B. In fact, this sequence's entropy per character in the A-optimal coding will be $-q \log_2 p - (1 - q) \log_2 (1 -p)$. This is the cross-entropy between A and B. The entropy per character of the sequence that B emits in its own optimal coding is $-q \log_2 q -(1 -q) \log_2 (1-q)$. The number of bits per character wasted to encode the sequence that B emits with the A-optimal coding is the relative entropy of A and B.

\begin{figure}[h!]
    \centerline{
    \includegraphics[width=0.35\linewidth]{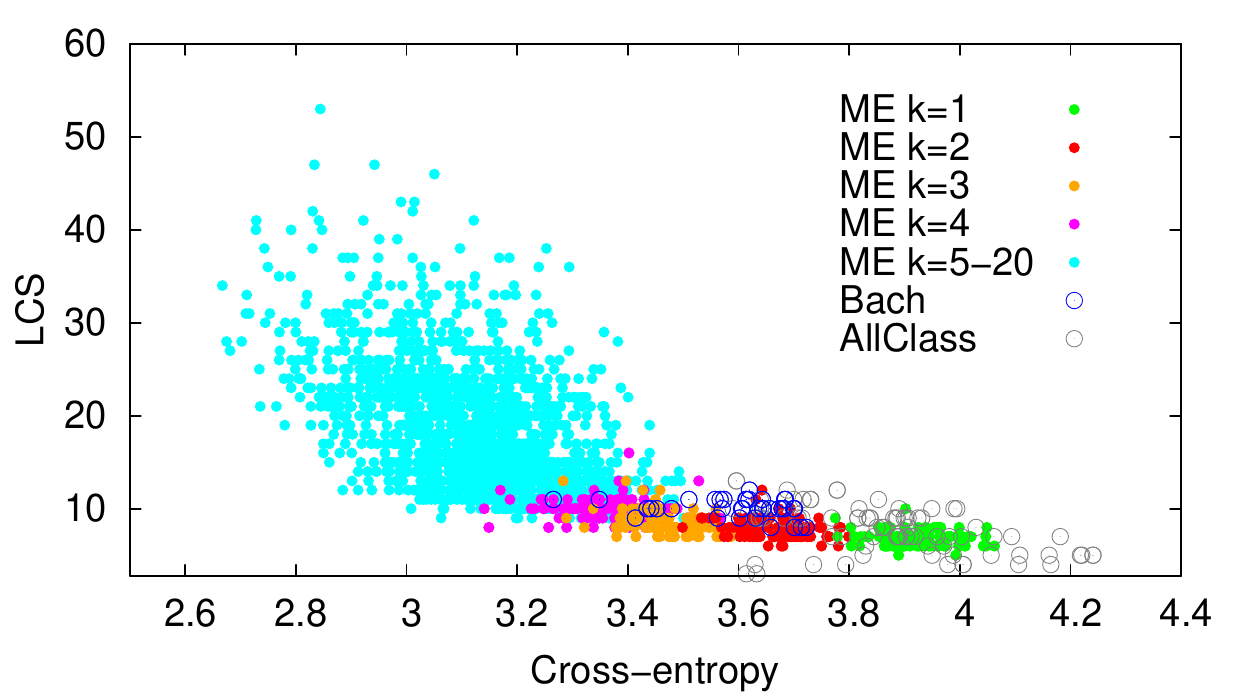}              
     \includegraphics[width=0.35\linewidth]{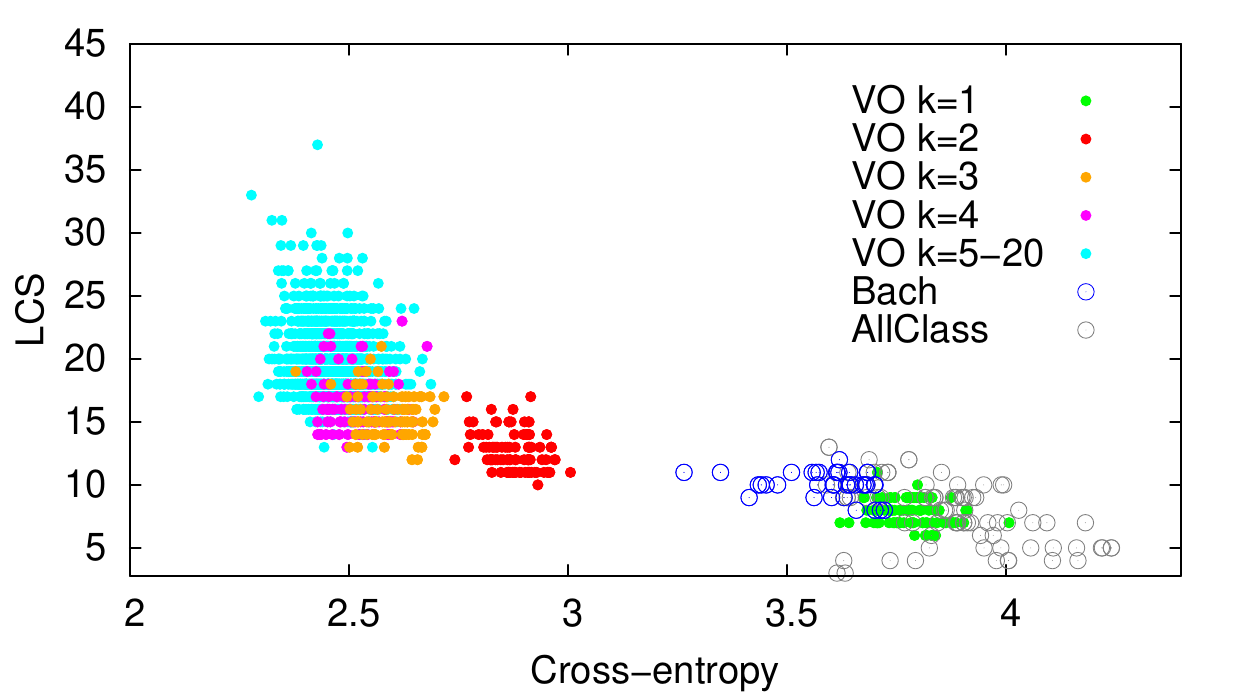}   
     \includegraphics[width=0.35\linewidth]{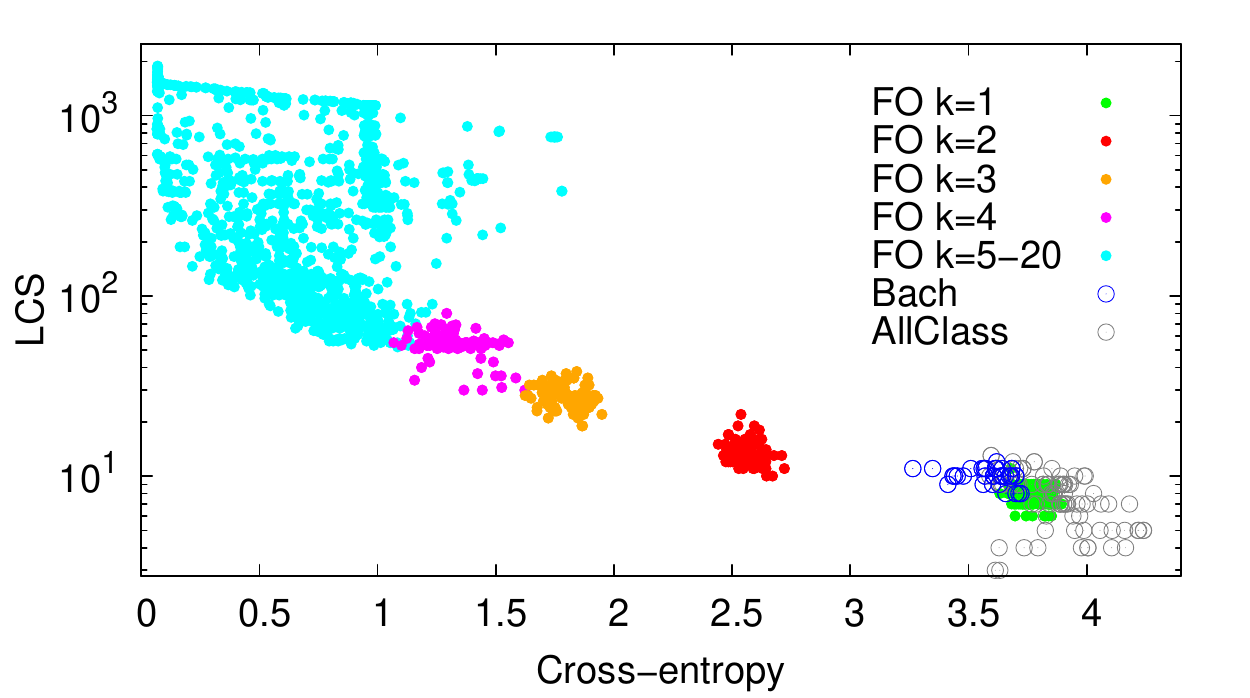}     }                    
    \caption{{\bf Borrowing vs. similarity.} This figure reports the Largest Common Substring (LCS) vs. the values of the cross-entropies for all the artificial sequences generated with the Maximum Entropy (ME) model (left), the variable-order (VO) Markov model (center) and the fixed-order (FO) Markov model (right). Everything is computed with respect to the sequence of J.S. Bach's first violin Partita (see Section S3 for details). Filled circles correspond to the artificial sequences. Colors code for the values of $k$ in each different model. In addition in each panel  the empty circles reports the same quantities for  other original sequences of Bach (represented with blue circles) and other classical authors (AllClass represented with grey circles). }
    \label{SIfig:Innovation_vs_similarity_LCS}
\end{figure}

The approach proposed in~\cite{bcl_2002,zippers_2005} estimates the cross-entropy between two sequences by using the LZ77 data compression scheme. The LZ77 algorithm first looks for duplicated strings in the input data. It replaces the second occurrence of a string with a pointer to the previous string. This pointer consists of two numbers: a distance, representing how far back into the window the sequence starts, and the length in characters of that subsequence. The original LZ77 algorithm defines the window as the section of the sequence already scanned sequentially. In our implementation of LZ77 we scan the sequence 
$\cal{B}$ by looking for matches only  in sequence $\cal{A}$. In this way the algorithm is automatically looking for the best encoding of sequence $\cal{B}$ using the best code for sequence $\cal{A}$.

\section*{S6 Robustness of the results}
\label{subsec:robustness}

In this section we report additional examples of the results presented in the section {\em Borrowing vs. Innovation} of the main text. We considers in particular more original corpora coming from Beethoven, Schumann, Chopin and Liszt to demonstrate the robustness of the results. In particular, we show robustness both for different pieces of the same author  (Fig.~\ref{SIfig:Innovation_vs_similarity_beethoven} presents the results for three different pieces by Ludwig Van Beethoven) and for pieces of different authors  (Fig.~\ref{SIfig:Innovation_vs_similarity_SCL} presents the results for pieces by Robert Schumann, Fr\'ed\'eric Chopin and Franz Liszt).

Finally we complement Figure~6 of the main text by reporting the same results where we substituted the Average Common Substring (ACS) with the Longest Common Substring (LCS).

%% \section*{S7 Audio files}
%% \label{S7_Audio}
%% TBD

%% \section*{Acknowledgments}
%% This research is conducted within the Flow Machines project which received funding from the European Research Council under the European Union's Seventh Framework Programme (FP/2007-2013) / ERC Grant Agreement n. 291156. 
%% FT and VL acknowledge partial support from the KREYON project, funded by the Templeton Foundation, Grant Agreement n. 51663, for the completion of this work while in Turin and in Rome.

\bibliographystyle{ieeetr}
 \bibliography{maxentmusic}

\end{document}